\documentclass[final]{cvpr}

\usepackage{times}
\usepackage{epsfig}
\usepackage{graphicx}
\usepackage{amsmath}
\usepackage{amssymb}

\usepackage[pagebackref=true,breaklinks=true,colorlinks,bookmarks=false]{hyperref}

\def\cvprPaperID{0040} % *** Enter the CVPR Paper ID here
\def\confYear{CVPR 2021}

\begin{document}

%%%%%%%%% TITLE
% \title{A Real-Scene Raw-RGB Benchmark for Reflection Removal}
% \title{Towards Real-World Single Image Reflection Removal: A New Benchmark}

\title{A Categorized Reflection Removal Dataset with Diverse Real-world Scenes}

% --- DATASET statistics ---: 
% triplet sets: 602 (vs. 1103)
% cropped triplet sets: 760 valid triplet

% \affil[1]{Department of Computer Science, \LaTeX\ University}
% \affil[2]{Department of Mechanical Engineering, \LaTeX\ University}

% \author{First Author\\
% Institution1
% {\tt\small firstauthor@i1.org}

% For a paper whose authors are all at the same institution,
% omit the following lines up until the closing ``}''.
% Additional authors and addresses can be added with ``\and'',
% just like the second author.
% To save space, use either the email address or home page, not both
% \and
% Second Author\\
% Institution2
% % {\tt\small secondauthor@i2.org}
% \and
% Second Author\\
% Institution2
% \and
% Second Author\\
% Institution2
% \and
% Second Author\\
% Institution2
% }
\author{
\begin{tabular}{c@{\hspace{1cm}}c@{\hspace{1cm}}c@{\hspace{1cm}}c}
Chenyang Lei$^{1*}$ & ~~~Xuhua Huang$^{1,2}$\thanks{Equal contribution}& ~~~Chenyang Qi$^{1}$ &~~~Yankun Zhao$^{1}$\\
\end{tabular}\\
\begin{tabular}{ccc}
Wenxiu Sun$^{3}$ & ~~~Qiong Yan$^{3}$& ~~~Qifeng Chen$^{1}$
\end{tabular}
\\
\\
\begin{tabular}{ccc}
$^{1}$HKUST&
$^{2}$CMU&
$^{3}$SenseTime\\
\end{tabular}
\\
}

\maketitle

%%%%%%%%% ABSTRACT
\begin{abstract}
%The performance of most reflection removal methods is decided by the similarity between training data and real-world data. 

Due to the lack of a large-scale reflection removal dataset with diverse real-world scenes, many existing reflection removal methods are trained on synthetic data plus a small amount of real-world data, which makes it difficult to evaluate the strengths or weaknesses of different reflection removal methods thoroughly. Furthermore, existing real-world benchmarks and datasets do not categorize image data based on the types and appearances of reflection (e.g., smoothness, intensity), making it hard to analyze reflection removal methods. Hence, we construct a new reflection removal dataset that is categorized, diverse, and real-world (CDR). A pipeline based on RAW data is used to capture perfectly aligned input images and transmission images. The dataset is constructed using diverse glass types under various environments to ensure diversity. By analyzing several reflection removal methods and conducting extensive experiments on our dataset, we show that state-of-the-art reflection removal methods generally perform well on blurry reflection but fail in obtaining satisfying performance on other types of real-world reflection. We believe our dataset can help develop novel methods to remove real-world reflection better. Our dataset is available at \href{https://alexzhao-hugga.github.io/Real-World-Reflection-Removal/}{https://alexzhao-hugga.github.io/Real-World-Reflection-Removal/}.

% named \textbf{CDR} that is an order of magnitude larger than existing real-world datasets for reflection removal. Our dataset consists of ground-truth reflection, transmission, and mixture images in both raw and RGB formats with diverse glass types and environments. By analyzing a diverse set of methods and conducting extensive experiments on our dataset, we show that state-of-the-art reflection removal algorithms are still far from perfect and some open challenges are to be solved.

% Although existing reflection removal methods appear to have high performance on both synthetic and limited controlled real data in public benchmark, we observe that their performance is still unsatisfactory on more general real-scene images. Arguably, the evaluation of reflection removal should be conducted on a large real-scene dataset captured by mimicking normal users. Hence, we construct a new real-scene dataset named \textbf{CDR} that is an order of magnitude larger than existing real-world datasets for reflection removal. Our dataset consists of ground-truth reflection, transmission, and mixture images in both raw and RGB formats with diverse glass types and environments. By analyzing a diverse set of methods and conducting extensive experiments on our dataset, we show that state-of-the-art reflection removal algorithms are still far from perfect and some open challenges are to be solved. %We analyze a diverse set of methods in different situations and discuss open challenges in the end.
\end{abstract}

%%%%%%%%% BODY TEXT

\begin{figure*}[t]
\centering
\begin{tabular}{@{}c@{\hspace{1mm}}c@{\hspace{1mm}}c@{\hspace{1mm}}c@{\hspace{1mm}}c@{\hspace{1mm}}c@{\hspace{1mm}}c@{}}
\rotatebox{90}{ \hspace{6mm} \small Blurry}&
\includegraphics[width=0.15\linewidth]{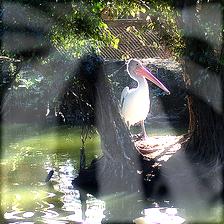}&
\includegraphics[width=0.15\linewidth]{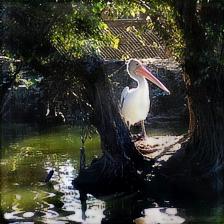}&
\includegraphics[width=0.15\linewidth]{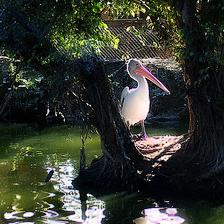}&
\includegraphics[width=0.15\linewidth]{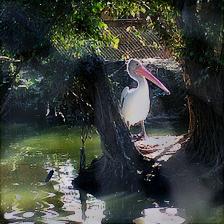}&
\includegraphics[width=0.15\linewidth]{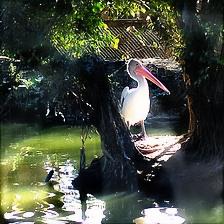}&
\includegraphics[width=0.15\linewidth]{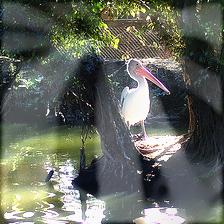}\\
\rotatebox{90}{ \hspace{4mm} \small Blurry}&
\includegraphics[width=0.15\linewidth]{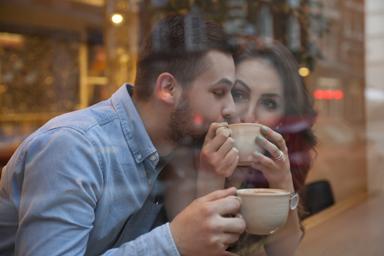}&
\includegraphics[width=0.15\linewidth]{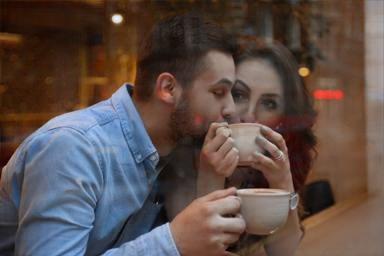}&
\includegraphics[width=0.15\linewidth]{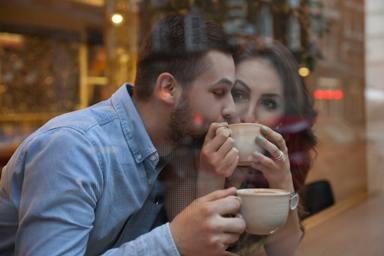}&
\includegraphics[width=0.15\linewidth]{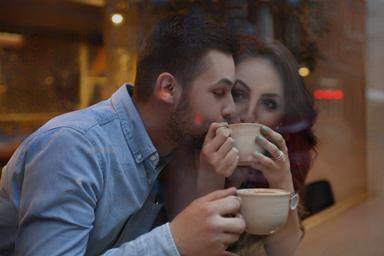}&
\includegraphics[width=0.15\linewidth]{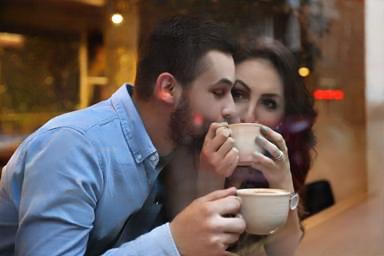}&
\includegraphics[width=0.15\linewidth]{Figure/Introduction/Fig1/ConReal/zhang.jpg}\\
\rotatebox{90}{\hspace{4mm} \small Sharp}&
\includegraphics[width=0.15\linewidth]{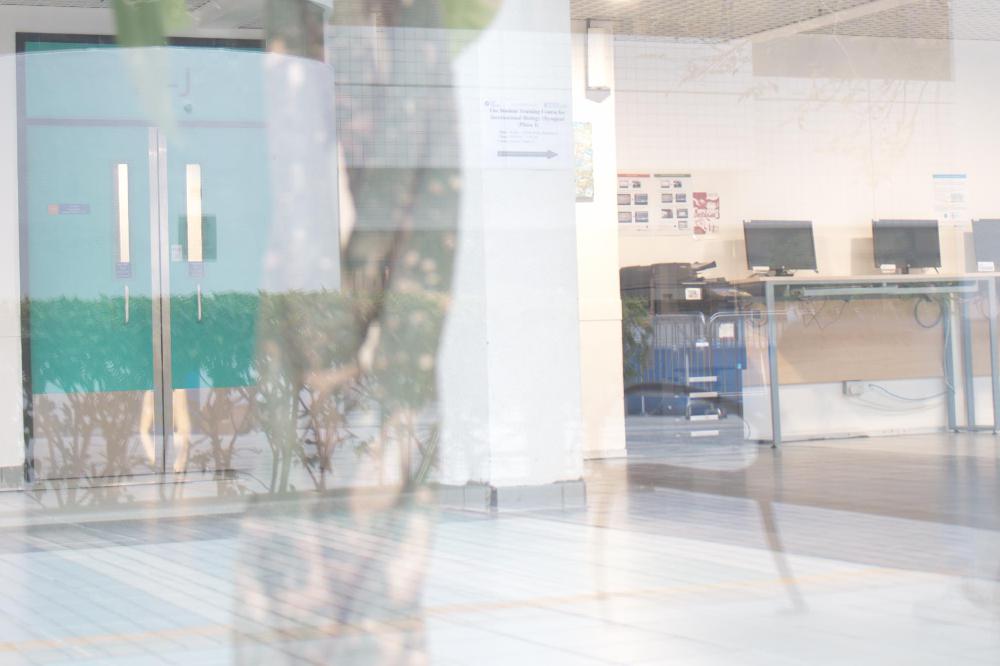}&
\includegraphics[width=0.15\linewidth]{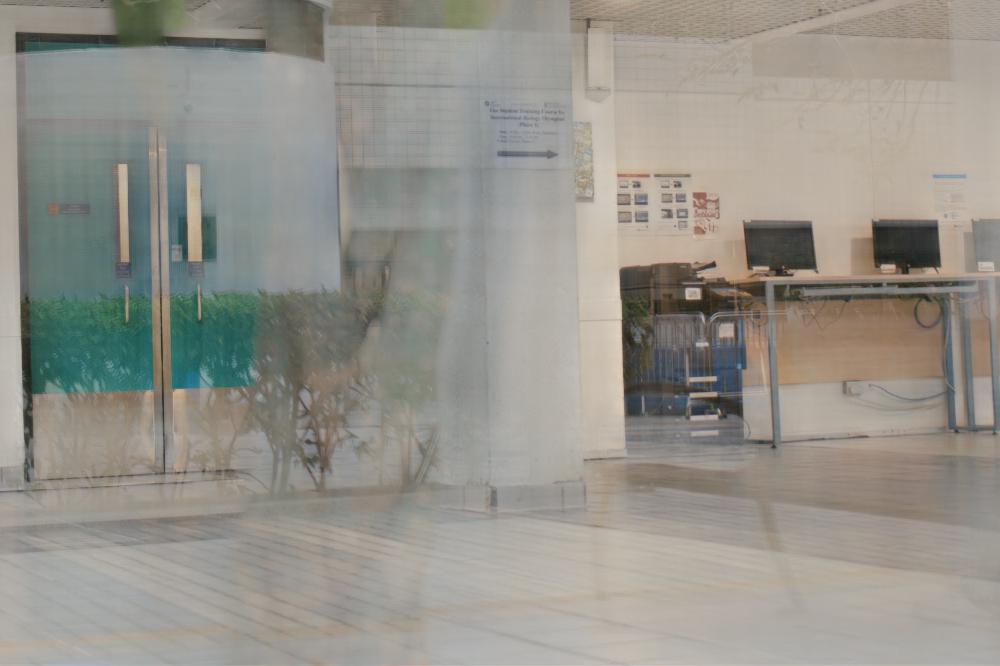}&
\includegraphics[width=0.15\linewidth]{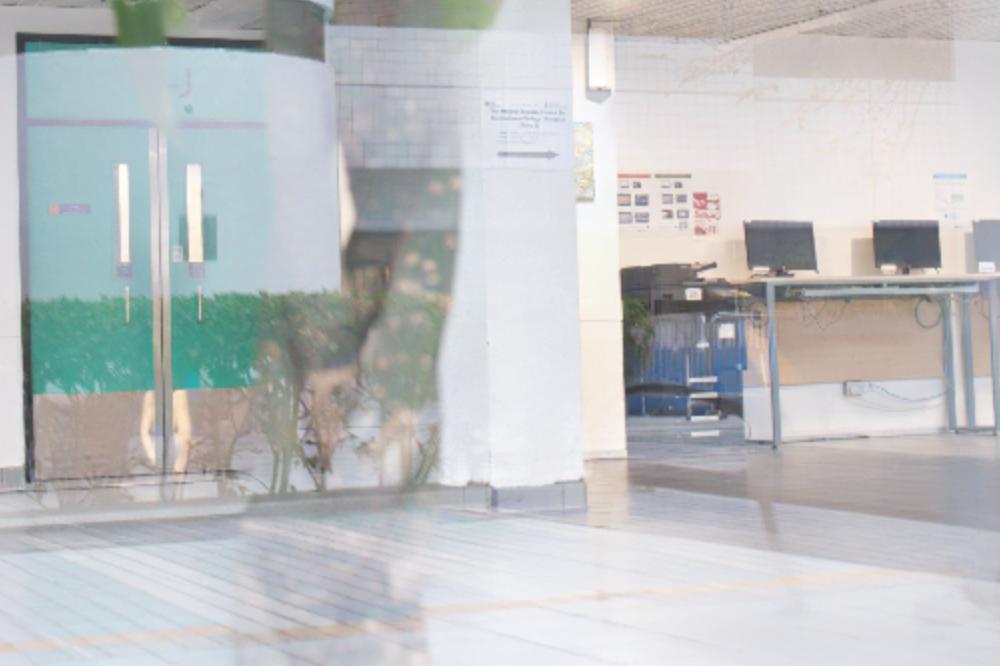}&
\includegraphics[width=0.15\linewidth]{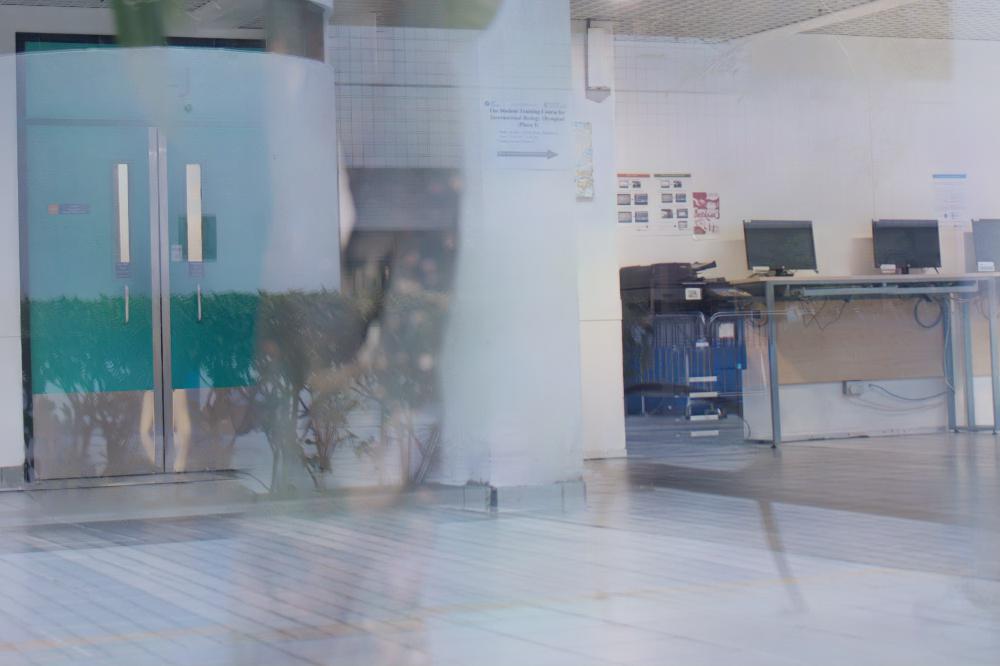}&
\includegraphics[width=0.15\linewidth]{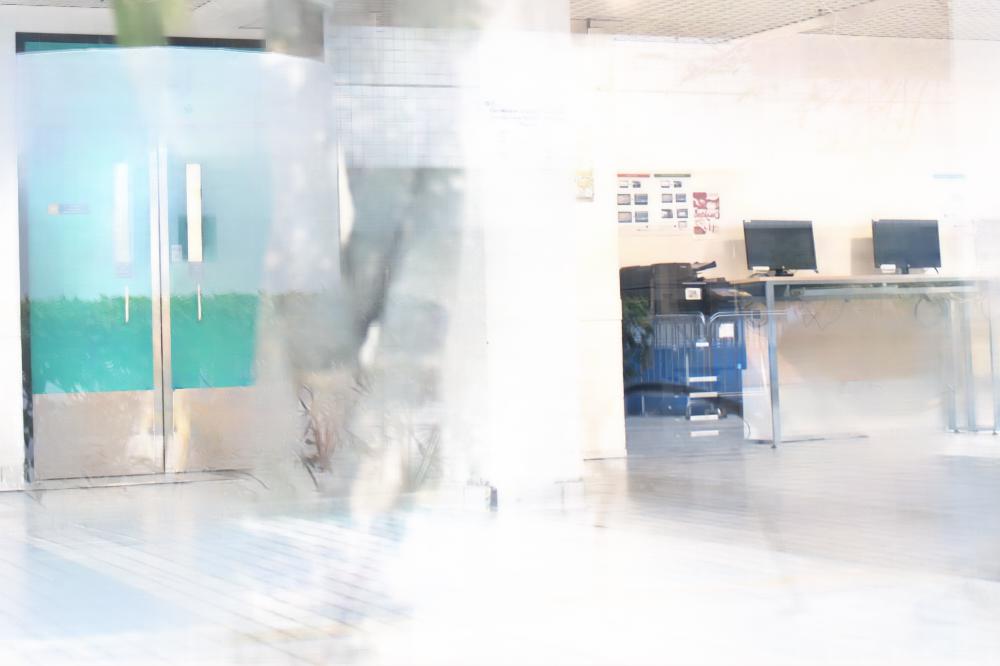}&
\includegraphics[width=0.15\linewidth]{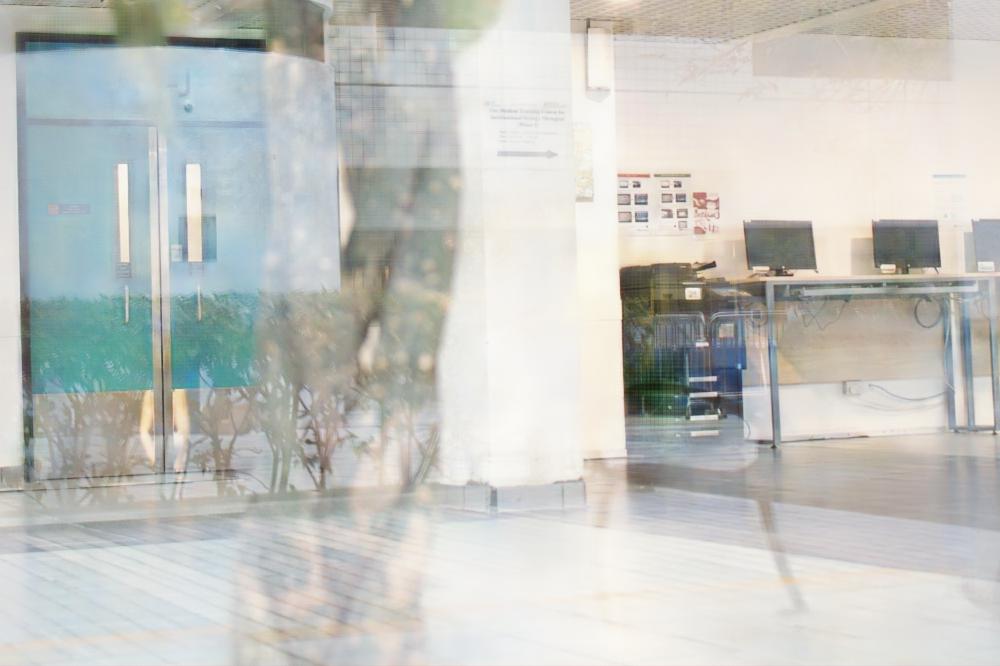}\\
\rotatebox{90}{\hspace{5mm} \small Sharp}&
\includegraphics[width=0.15\linewidth]{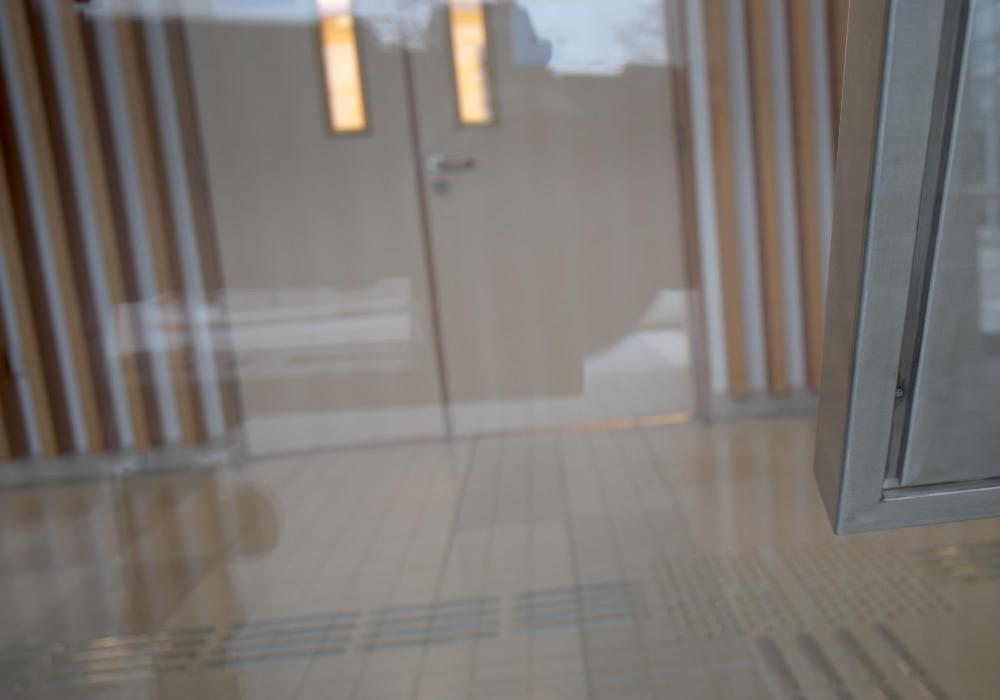}&
\includegraphics[width=0.15\linewidth]{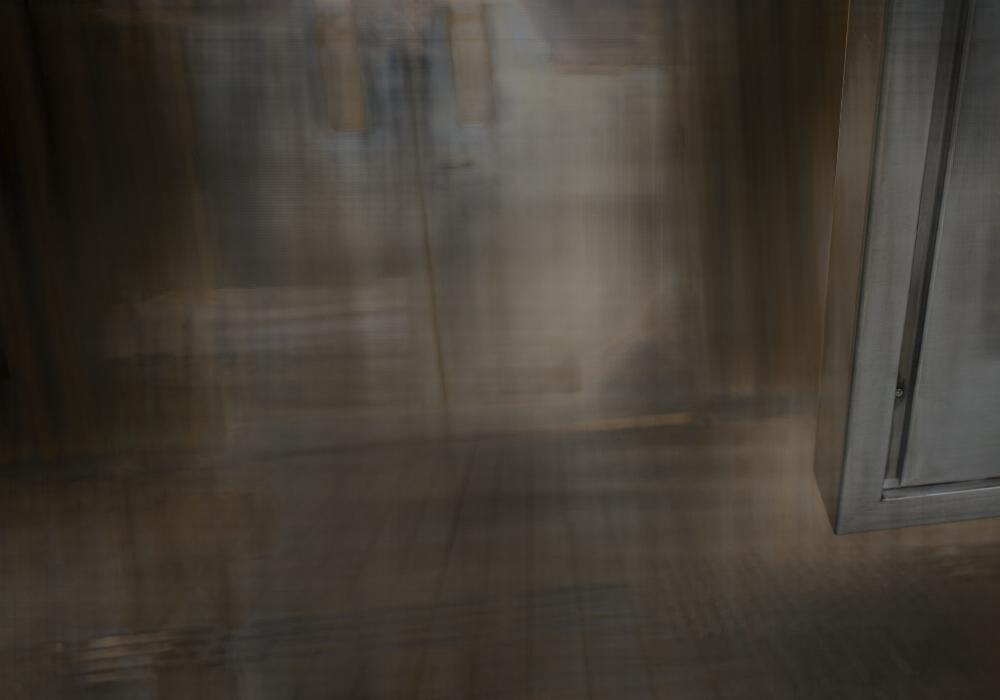}&
\includegraphics[width=0.15\linewidth]{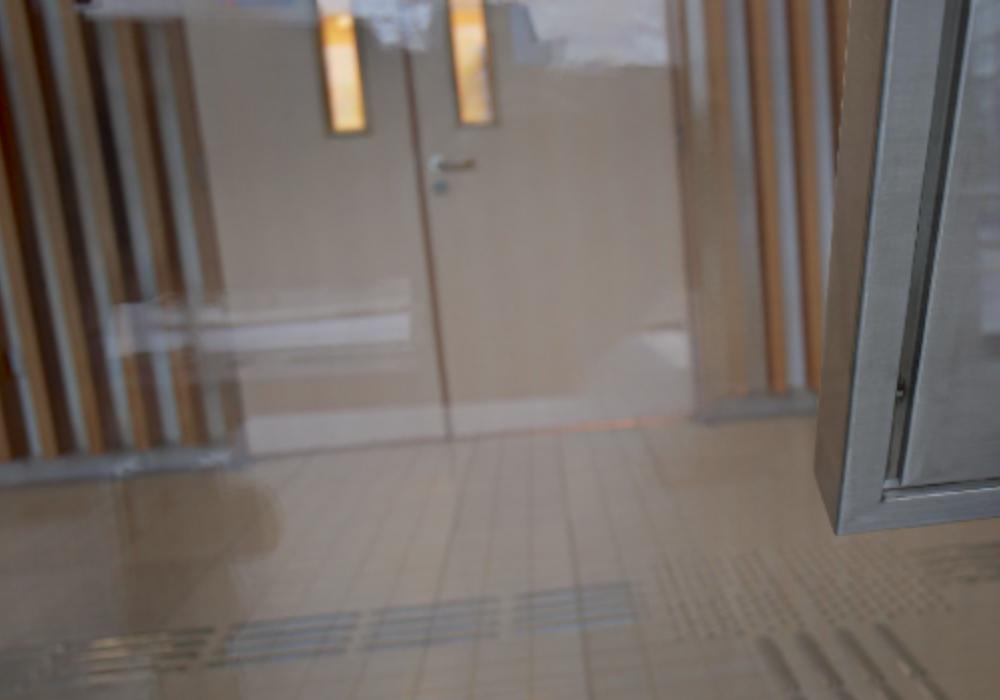}&
\includegraphics[width=0.15\linewidth]{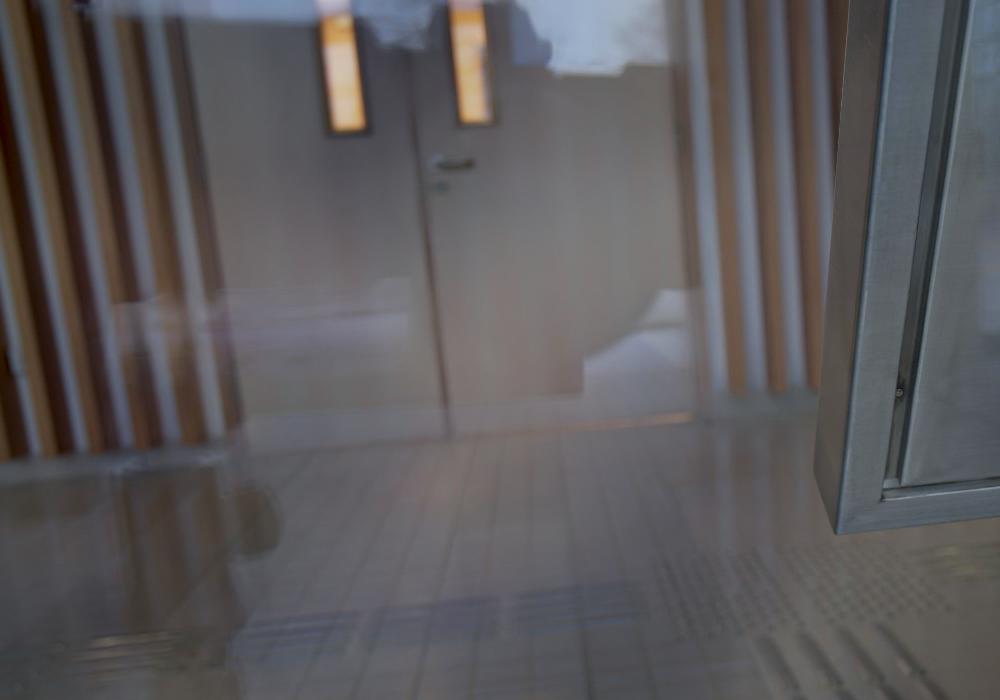}&
\includegraphics[width=0.15\linewidth]{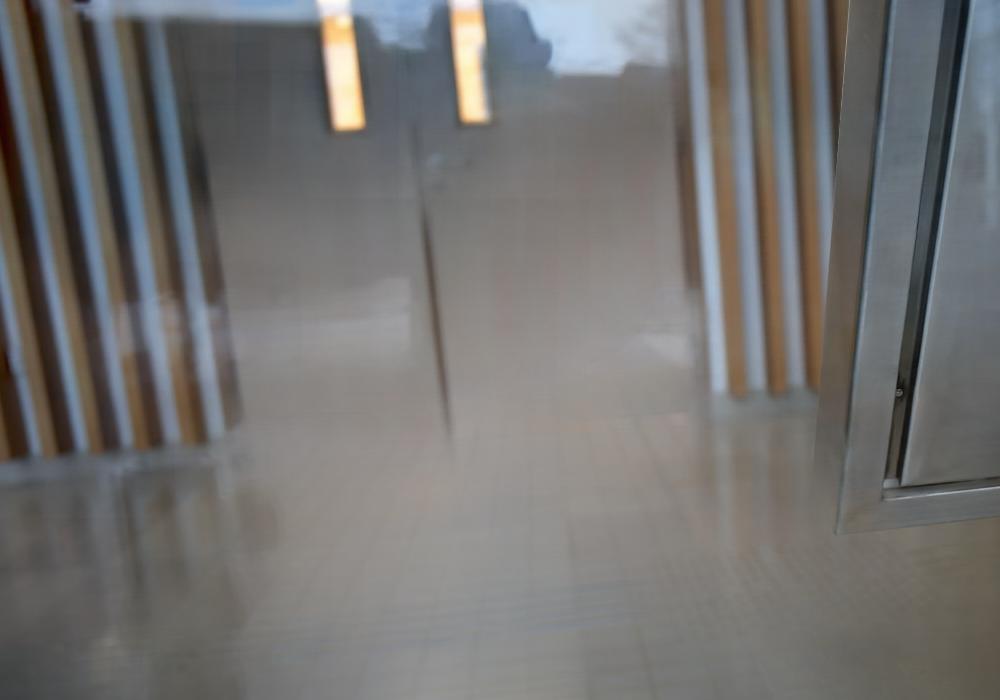}&
\includegraphics[width=0.15\linewidth]{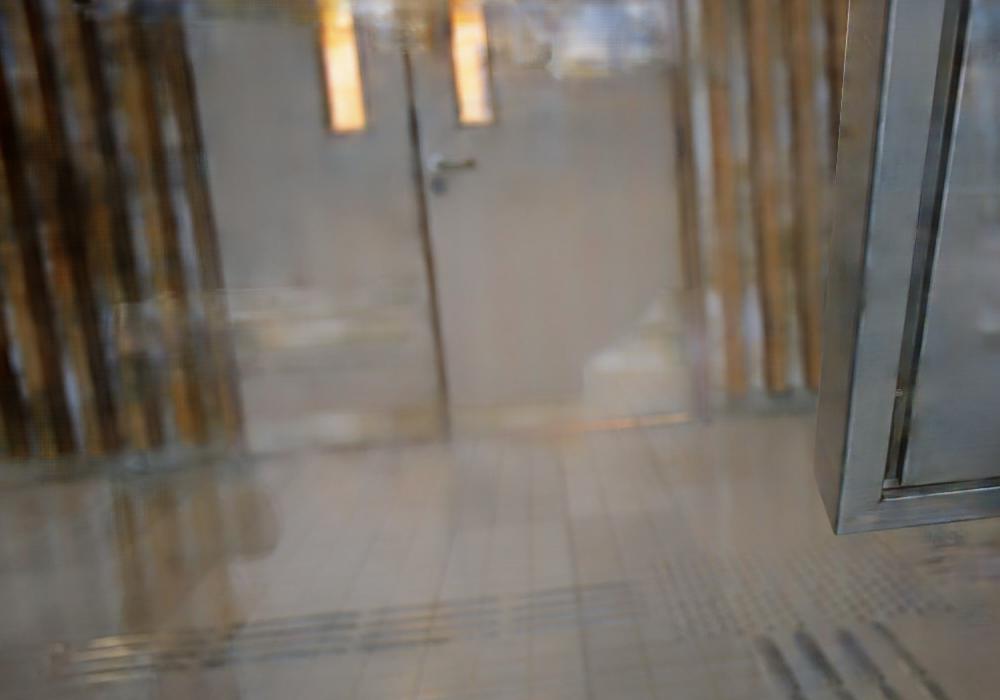}\\

&\small{Input} & 
\small{Zhang et al.~\cite{zhang2018single}} & \small{Wei et al.~\cite{wei2019single_ERR}} & 
\small{CoRRN~\cite{CoRRN}} & 
\small{BDN~\cite{eccv18refrmv_BDN}} & \small{Wen et al.~\cite{Wen_2019_CVPR_Linear}} \\
\end{tabular}
\caption{The performance of existing methods on different types of reflection is quite different. Most algorithms can remove the blurry reflection but cannot remove the sharp reflection well.}%We raise a question in the process of building this benchmark: if the photos are taken by normal users, can these state-of-the-art methods perform well? We collect the data carefully, but most algorithms still fail in many cases. The first row, second row, and the last two rows present results on synthetic data, controlled real data, and general real-scene data, respectively. Controlled real data often mimics the assumption of synthetic data.
\label{fig:BadPerformance}
\end{figure*}
\vspace{-5mm}

\section{Introduction}
\label{sec:intro}

Reflection removal is a task to remove undesirable reflection artifacts from a photograph. Existing deep learning based approaches for reflection removal have demonstrated superior performance on synthetic data, but we find that their performance degrades severely in diverse real-world data, as shown in Fig.~\ref{fig:BadPerformance}. 
% We conjecture that this is because the data used in prior work are not enough to represent general real-world scenes. These data include synthetic data, real-world data in lab environments, and a limited number of real-scene images. 
Most existing learning-based methods~\cite{fan2017generic,zhang2018single} are trained on synthetic data created under various assumptions. Hence, their performance is limited by the domain gap between real-world and synthetic data. The assumptions to create the synthetic data are often simplified, and thus these approaches have sub-optimal performance on real-world data.

The existing real-world benchmark dataset SIR$^2$ facilitates the research in reflection removal, but it has weaknesses as reported in the paper~\cite{wan2017benchmarking}. In this dataset, most images only contain flat objects or small object (i.e., postcards and solid objects) under controlled lighting. These images do not represent the scenes in our daily live because object distance, scales, and natural illumination variation are quite diverse in the wild. In the SIR$^2$ dataset, only collected 55 pairs of images with ground-truth transmission in the wild. There are also other real-world datasets~\cite{zhang2018single,Li_2020_CVPR} with high-quality ground truth but the they have a small number of images.

% most samples in SIR$^2$ are controlled real data and are different with general real scene data: depth, distance, scales, and natural illumination variation are quite diverse in real scene while their controlled real data only contain flat objects (i.e., postcards) or similar scales (i.e., solid objects) under controlled light. Though they collected 55 pairs of images in wild scene as a supplement, the other 400 pairs of images are still from controlled scene images. %We observe that this phenomenon is common: while the performance of algorithms on the SIR$^2$ benchmark is competitive, the performance of general real scene is not as good as the results shown on SIR$^2$.  

%A better real-world benchmark with high quality and scene diversity is needed to facilitate the research in reflection removal. 
%A reflection removal algorithm aims to estimate transmission T from an input image $M$ since reflection will have a noticeable bad influence on the quality of photos or the performance of computer vision tasks, such as object detection, image classification and image matching.

To address the limitations of existing reflection removal datasets, we present a large reflection removal dataset CDR that contains diverse scene in the real world. Compared with prior work, CDR has several advantages, as shown in Table~\ref{table:Dataset comparison}. In terms of the number of real-world images, ours is much larger and more diverse than existing ones. We construct the dataset following several principles to ensure image quality and diversity. First, we capture our data in the wild since it is similar to images captured in daily life. Second, to ensure perfect alignment between transmission and input mixed images, we obtain the transmission by subtracting reflection from the mixed image in the raw data space~\cite{Lei_2020_CVPR}. Third, we capture the images using different glasses in diverse scenes to ensure the diversity of our dataset. %We also keep the raw data for further study.
%, as reported by SIR$^2$ and other work~\cite{wan2017benchmarking,Lei_2020_CVPR}
We believe that the performance of a reflection removal method is related to the smoothness of reflection, and thus we carefully categorize the collected data into different types of reflection. We split the data based on the smoothness of reflection (i.e., sharp reflection or blurry reflection). In our experiments,  all existing methods are sensitive to the smoothness of reflection. %Therfore, we believe more types of reflection should be explored beyond blurry reflection.

We hope the CDR dataset can facilitate the research in reflection removal. The CDR dataset can provide a more extensive evaluation for reflection removal methods. In addition, the detailed categorization can help analyze the bottlenecks of existing methods.

\section{Background}

%In this review, we categorize existing methods according to the number of input images and focus more on analyzing the assumptions of these methods. We will review the contribution and limitation of existing methods, datasets and benchmarks for reflection removal task.
%For example, we can classify by the cues each method reply on (e.g., gradient~\cite{Arvanitopoulos_2017_CVPR,Yang_2019_CVPR}, ghosting effect~\cite{shih2015reflection}), the nature of the method (e.g. deep-learning or not), or by the number of input images (e.g., single image, multiple images).

\subsection{Single Image Reflection Removal}
Most single image reflection removal methods~\cite{fan2017generic, zhang2018single, eccv18refrmv_BDN, Yang_2019_CVPR} rely on various assumptions. Considering image gradients, Arvanitopoulos et al.~\cite{Arvanitopoulos_2017_CVPR} propose the idea of suppressing the reflection, and Yang et al.~\cite{Yang_2019_CVPR} propose a faster method based on convex optimization. These methods fail to remove sharp reflection. Under the assumption that transmission is in focus, Punnappurath et al.~\cite{Punnappurath_2019_CVPR} design a method based on dual-pixel camera input. 
CEILNet~\cite{fan2017generic}, Zhang et al.~\cite{zhang2018single}, and BDN~\cite{eccv18refrmv_BDN} assume that reflection is out of focus and synthesize images to train their neural networks. CEILNet~\cite{fan2017generic} estimates target edges first and uses it as guidance to predict the transmission layer. Zhang et al.~\cite{zhang2018single} use perceptual and adversarial losses to capture the difference between reflection and transmission. BDN~\cite{eccv18refrmv_BDN} estimates the reflection images, which is then used to estimate the transmission layer. These methods~\cite{zhang2018single,fan2017generic,eccv18refrmv_BDN} work well when reflection is more defocused than transmission but fail otherwise. For these deep learning based approaches, training data is critical for good performance.
To bridge the gap between synthetic and real-world data, Zhang et al.~\cite{zhang2018single} and Wei et al.~\cite{wei2019single_ERR} collected some real-world images for training. However, their images have misalignment issues between transmission and input images, and the dataset size is small. Wei et al.~\cite{wei2019single_ERR} propose to use high-level features that are less sensitive to small misalignment to calculate losses.
To obtain more realistic and diverse data, Wen et al.~\cite{Wen_2019_CVPR_Linear} and Ma et al.~\cite{Ma_2019_ICCV} propose to synthesize data using a deep neural network and achieve better performance and generalization. Kim et al.~\cite{Kim_2020_CVPR} propose a physics based method to render the reflection and achieve better performance than using synthetic images.

\begin{table*}[t]
\centering
\renewcommand{\arraystretch}{1.2}
\begin{tabular*}{\textwidth}{@{}c@{\hspace{2mm}}c@{\hspace{2mm}}c@{\hspace{2mm}}c@{\hspace{2mm}}c@{\hspace{2mm}}c@{\hspace{2mm}}c@{\hspace{2mm}}c@{\hspace{2mm}}c@{}}
\hline
    & \small{Glass type} & \small{Categorized} & \small{Scene-level data} & \small{Alignment} & \small{Reflection image} & \small{Curve glass}& \small{Data type}& \small{Training set} \\ 

\hline
\small{SIR$^2$-Wild~\cite{wan2017benchmarking}}   & \small{3} & \small{No} & \small{55} & \small{Calibrated} & \small{Yes} & \small{No}& \small{RGB} & \small{No}\\ 
\small{Zhang et al.~\cite{zhang2018single}}   & \small{1} & \small{No} & \small{$<$110} & \small{Calibrated} & \small{No} & \small{No}& \small{RGB} & \small{No}\\ 
\small{Nature~\cite{Li_2020_CVPR}}   & \small{2} & \small{No} & \small{$<$ 220} & \small{Misalignment} & \small{No} & \small{No}& \small{RGB} & \small{No}\\ 
\small{Ours}    & \small{\textgreater 200 } & \small{Yes} & \small{1,063} & \small{Perfect} & \small{Yes} & \small{Yes} & \small{RGB\&Raw} &\small{Yes}\\ 
\hline
\end{tabular*}
\vspace{1mm}
\caption{The comparisons between our data and existing datasets~\cite{wan2017benchmarking,zhang2018single,Li_2020_CVPR}. 
Scene-level data: the data that is captured in the wild instead of lab environments. } %We capture large-scale dataset in the wild with perfect alignment. In addition, our dataset contains reflection, raw data and training set.The number of the triple set (i.e., data with ground-truth) is presented in parentheses.
\label{table:Dataset comparison}

\end{table*}

\subsection{Reflection Removal with Multiple Images }
Utilizing multiple images as input provides additional information, which makes it possible to relax some strict assumptions used in prior work. A number of approaches~\cite{szeliski2000layer,Sarel2005,sarel2004separating,li2013exploiting,guo2014robust,han2017reflection,xue2015computational,sun2016automatic,alayrac2019visual,Liu_CVPR_2020} exploit the relative motion between reflection and transmission with multiple images captured with camera movement to remove reflection.

Some other methods may take a sequence of images under specific conditions or camera settings. For example, pairs of flash and no-flash images~\cite{DBLP:journals/tog/AgrawalRNL05,Lei_2021_RFC}, near-infrared cameras~\cite{NIR_Hong},  light field cameras ~\cite{wang2015automatic}, dual-pixel cameras~\cite{Punnappurath_2019_CVPR}, and polarization cameras ~\cite{Lyu_2019_Polar,eccv2018/Wieschollek,2000Schechner,kong14pami,Fraid1999} can be used.

Different from supervised learning models, Double-DIP~\cite{DoubleDIP} separates a mixed image into reflection and transmission layers based on internal self-similarities in multiple superpositions, in an unsupervised fashion.

Although the methods based on multiple images do not rely on strict assumptions for appearances of reflections (e.g., blurry reflection, ghosting effects), they need additional requirements on data~\cite{xue2015computational} or special devices~\cite{Lei_2020_CVPR}, which may prevent them from broader applications. Therefore, in this work, we focus on building a dataset and evaluating single image reflection removal methods.

\subsection{SIR$^2$ benchmark dataset}
Wan et al.~\cite{wan2017benchmarking} propose the SIR$^2$ benchmark dataset for single image reflection removal. This dataset contains images taken in controlled scenes and in the wild. To solve the misalignment problem, they calibrate the alignment between the mixed image $M$ and the background $B$. The dataset was captured with three glasses of different thicknesses, various combinations of aperture sizes, and different exposure times to improve the diversity in this dataset. %However, as introduced in Sec.~\ref{sec:intro}, their benchmark has limitations.

As described by Wan et al.~\cite{wan2017benchmarking}, most of the objects in their controlled scene contain only flat objects (postcard) or objects with similar scales (solid objects). However, real-world scenes contains objects at different depths, and the natural environment illumination also varies greatly, while the controlled scenes are mostly captured in an indoor office environment. To address this limitation, 55 pairs of images with ground-truth reflection and transmission are captured in the wild, but 55 pairs are far from large scale. Also, this dataset does not provide a standard split among the training set, the validation set, and the test set.

%------------------------------------------------------------------------
\section{CDR Dataset}

In this section, we describe the features of our CDR dataset for reflection removal, which stands for ``Categorized, Diverse, and Real-world.'' A triplet $\{M, R, T\}$ is collected in each scene where $M$ is the mixed image, $R$ is the reflection image, and $T$ is the transmission image.

\begin{figure}[t]
\centering
\includegraphics[width=1\linewidth]{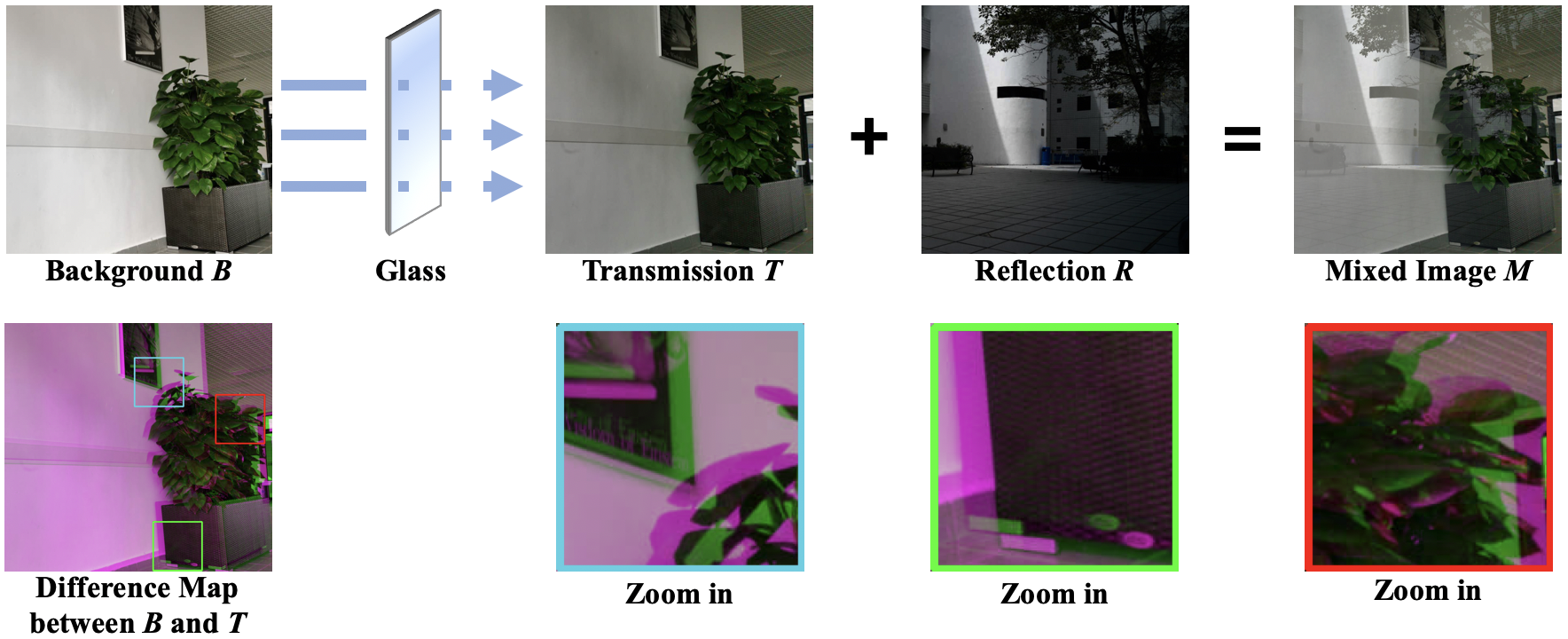}
\caption{Due to refraction, spatial shift and intensity difference exist between $B$ and $T$. The difference map visualizes the misalignment between $B$ and $T$. The sum of the reflection $R$ and the transmission $T$ equals to the mixed image $M$ in the raw data space.}
\label{fig:intro_MMR}
\end{figure}
% \vspace{5mm}

% \begin{figure*}[t]
% \centering
% \begin{tabular}{@{}c@{\hspace{1mm}}c@{\hspace{1mm}}c@{\hspace{1mm}}c@{\hspace{1mm}}c@{}}

% % \rotatebox{90}{\small \hspace{4mm}After ISP}
% \includegraphics[width=0.8\linewidth]{Figure/Method/samples.png}
% % &\includegraphics[width=0.24\linewidth]{Figure/Method/sample1.png}&
% % \includegraphics[width=0.24\linewidth]{Figure/Method/sample2.png}&
% % \includegraphics[width=0.24\linewidth]{Figure/Method/sample3.png}&
% % \includegraphics[width=0.24\linewidth]{Figure/Method/sample4.png}\\
% % &Example 1 & Example 2 & Example 3 & Example 4 \\
% \end{tabular}
% \caption{Some examples of our dataset.}
% \label{fig:MMR_sample}
% \end{figure*}

\begin{figure*}[t]
\centering
\begin{tabular}{@{}c@{\hspace{1mm}}c@{\hspace{1mm}}c@{\hspace{3mm}}c@{\hspace{3mm}}c@{\hspace{1mm}}c@{\hspace{1mm}}c@{}}

\includegraphics[width=0.15\linewidth]{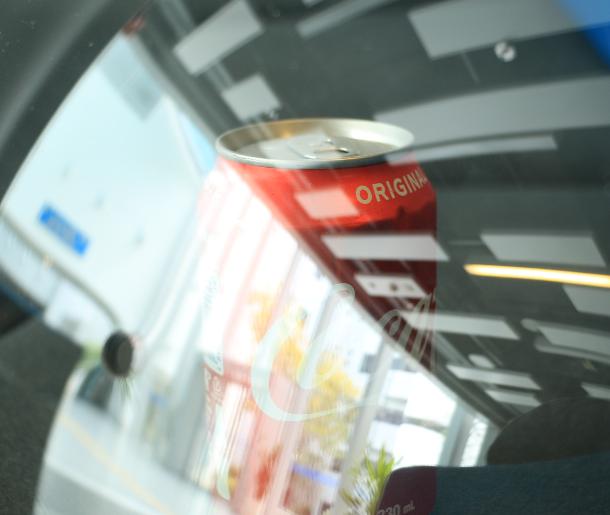}&
\includegraphics[width=0.15\linewidth]{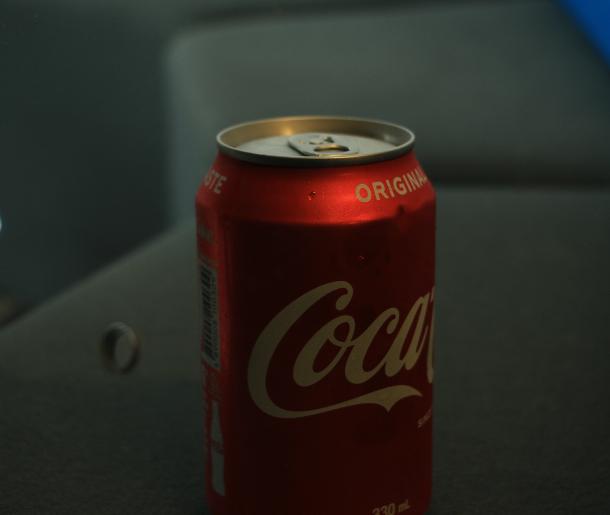}&
\includegraphics[width=0.15\linewidth]{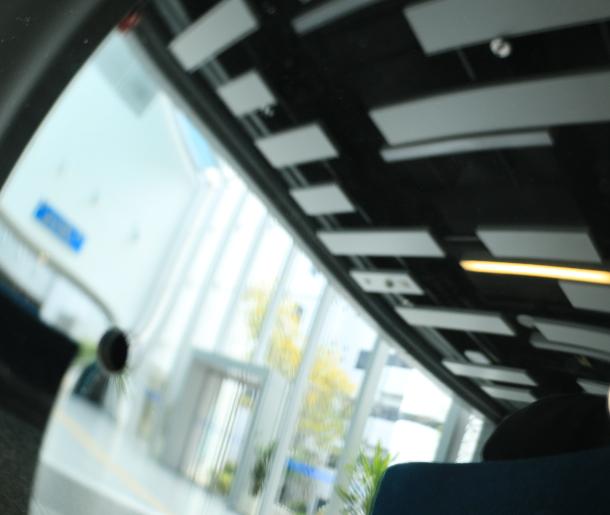}&
&
\includegraphics[width=0.15\linewidth]{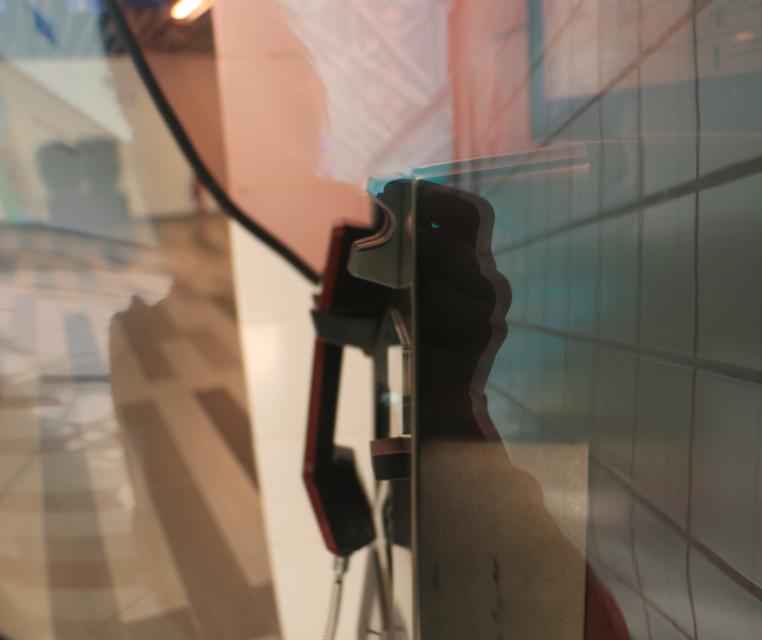}&
\includegraphics[width=0.15\linewidth]{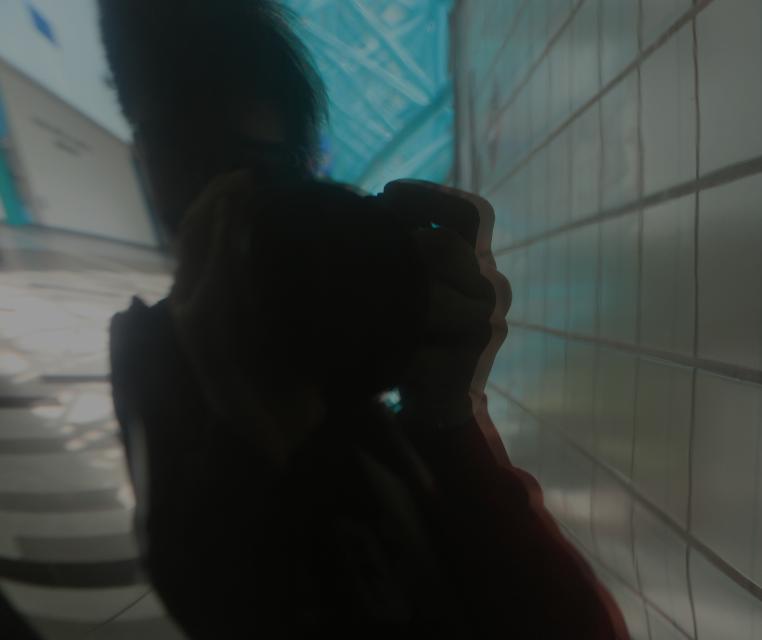}&
\includegraphics[width=0.15\linewidth]{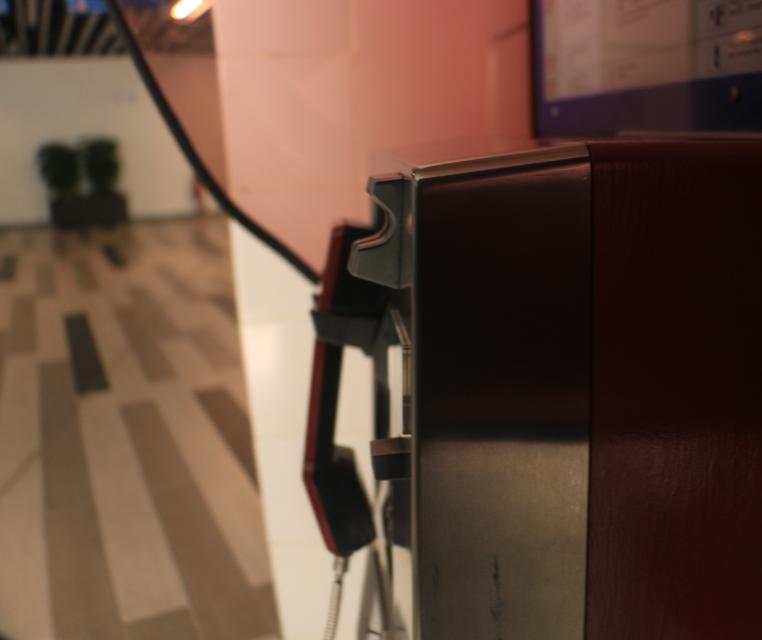}\\

\multicolumn{3}{c}{\{$M$,$R$,$T$\} on curved glass} &  & \multicolumn{3}{c}{\{$M$,$R$,$T$\} on colored (red) glass} \\

\includegraphics[width=0.15\linewidth]{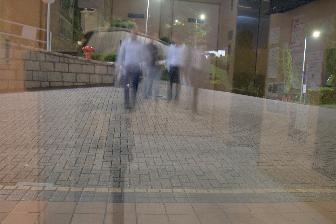}&
\includegraphics[width=0.15\linewidth]{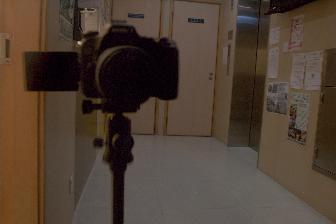}&
\includegraphics[width=0.15\linewidth]{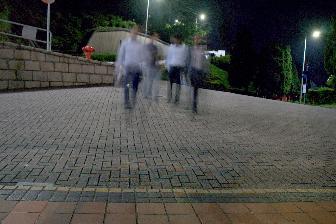}&
&
\includegraphics[width=0.15\linewidth]{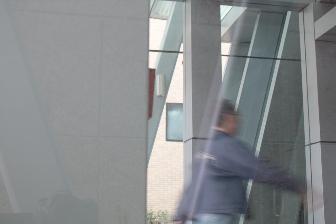}&
\includegraphics[width=0.15\linewidth]{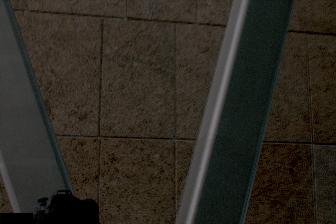}&
\includegraphics[width=0.15\linewidth]{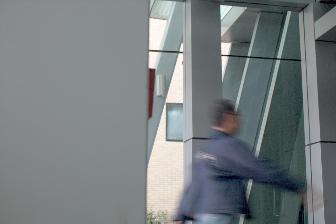}\\

\multicolumn{3}{c}{\{$M$,$R$,$T$\} on dynamic transmission} & & \multicolumn{3}{c}{\{$M$,$R$,$T$\} on dynamic transmission} \\

\end{tabular}
% \caption{We capture the data with colored and curved glass in dynamic transmission.}
\caption{More examples about the data diversity. In addition to glass types, we are also able to capture dynamic scenes, which enriches the scene diversity.}
\label{fig:DiverseData}
\end{figure*}
\begin{figure}[t]
\centering
\begin{tabular}{@{}c@{\hspace{1mm}}c@{\hspace{1mm}}c@{\hspace{1mm}}c@{}}
\includegraphics[width=0.24\linewidth]{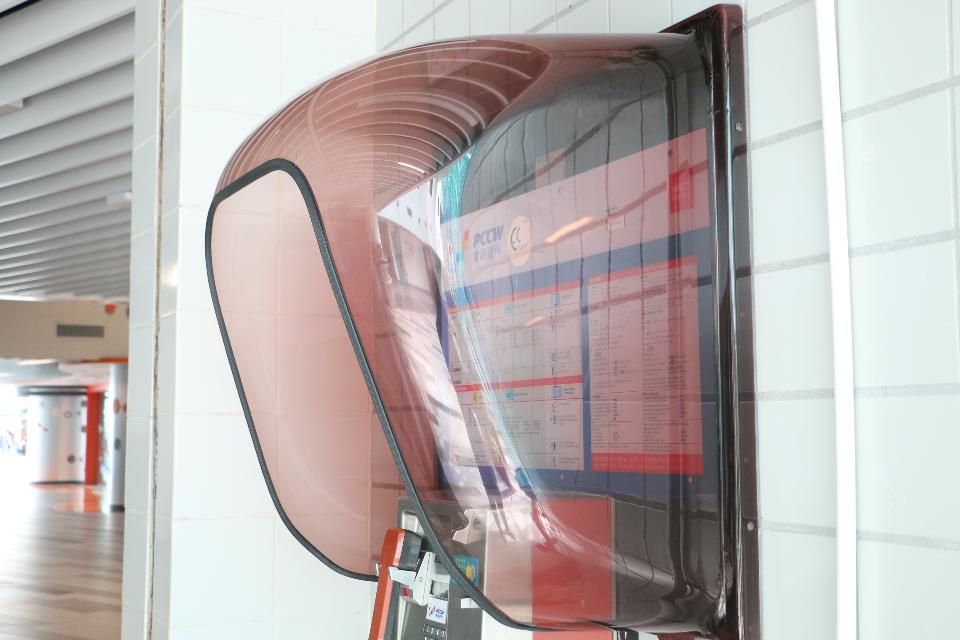}&
\includegraphics[width=0.24\linewidth]{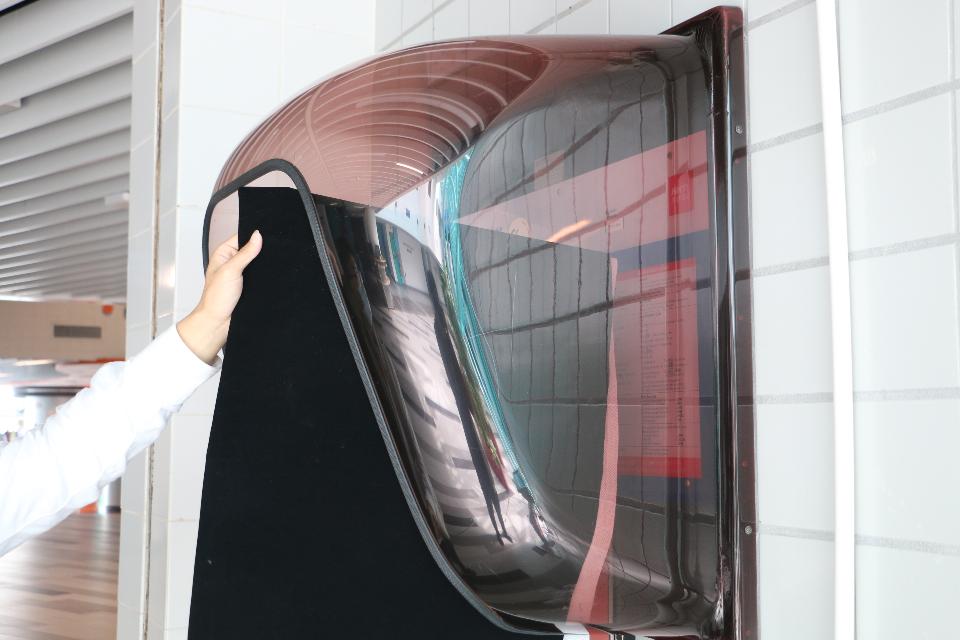}&
\includegraphics[width=0.24\linewidth]{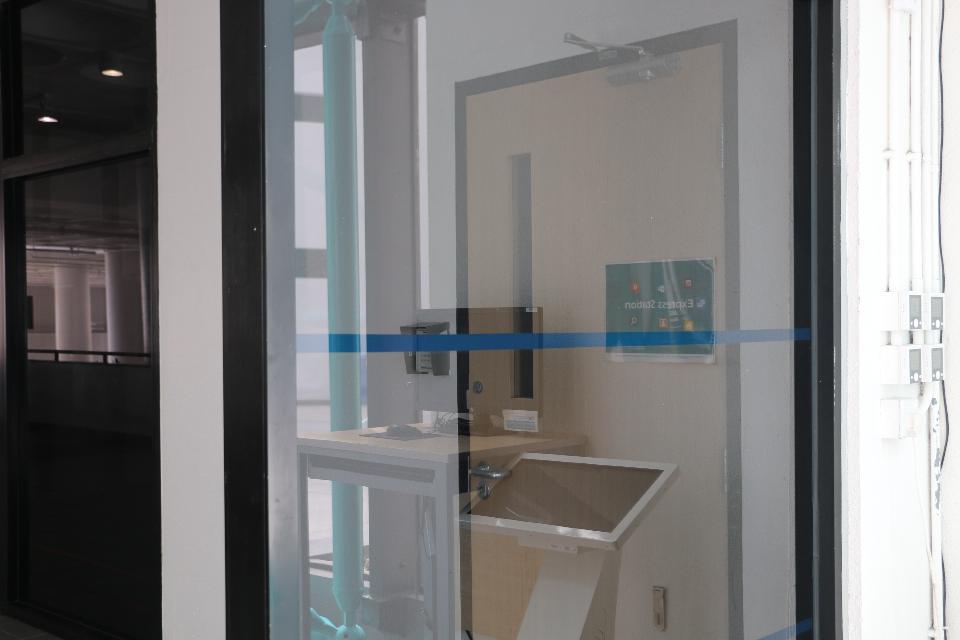}&
\includegraphics[width=0.24\linewidth]{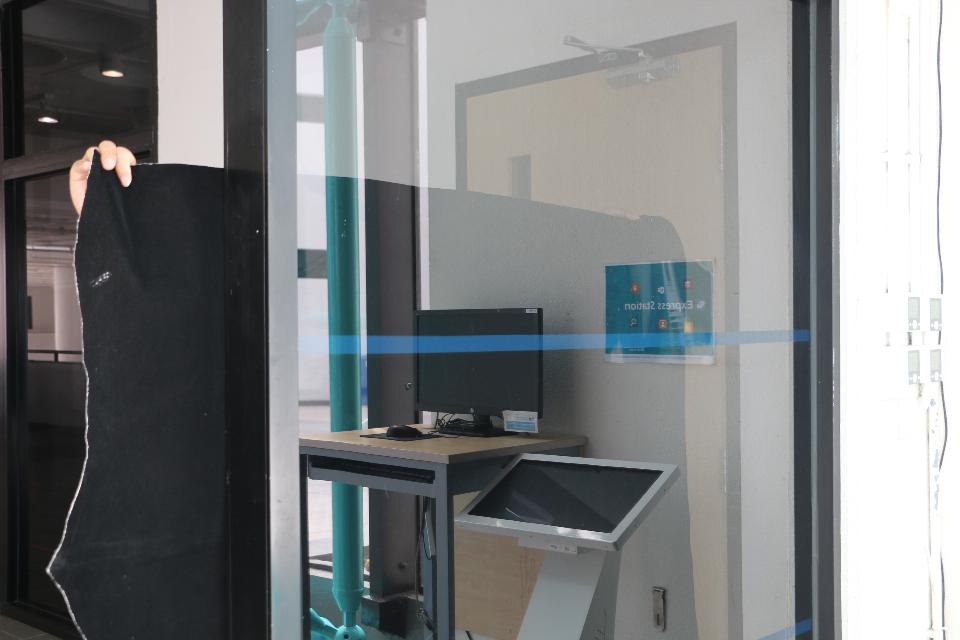}\\
Ours-$M1$ & Ours-$R1$& Ours-$M2$ & Ours-$R2$\\
\end{tabular}
\caption{With the M-R pipeline proposed by Lei et al.~\cite{Lei_2020_CVPR}, we can utilize a diverse set of glasses existing in our daily life (e.g., the curved and colored glass on the telephone booth, and the glass as a door).}
%types, including glass that are colored, curved, and thick. 

\label{fig:DataCollection}
\end{figure}

\label{subsec:data_description}
This dataset is collected mainly by three cameras: a DSLR (Digital Single-Lens Reflex) camera Canon EOS 50D, a MILC (Mirrorless Interchangeable Lens Cameras) Nikon Z6, and a smartphone camera Huawei Mate30. In total, we provide 1,063 triplets in our dataset.

We collected the real-world data in the wild. Compared with data captured in a controlled environment, real-wold data in the wild contains objects of various distances and illumination variation, as reported by SIR$^2$~\cite{wan2017benchmarking}. %Previous work collects limited number of real-world data in the wild (e.g., 55 pairs for SIR$^2$). %Images collected in the wild are more close to the images taken by normal users, which makes it better to evaluate the practicality of algorithms. In SIR$^2$, only 55 pairs of images with ground truth in the wild are provided. 
We improve our dataset in various aspects. Table~\ref{table:Dataset comparison} summarizes the main differences between the CDR dataset and existing datasets~\cite{wan2017benchmarking,zhang2018single,Li_2020_CVPR}. Specifically, our main advantages are listed as followed:

\noindent \textbf{(a) Data categorization.} We notice that the performance of a reflection removal model is related to the appearance of reflection. To facilitate in-depth analysis, we split all the images according to the reflection types. %Detailed analysis is presented in the experiments.%: relative intensity, smoothness between $R$ and $T$ and the ghosting effect. In our experiments, we notice existing methods demonstrate different performance on various types of data.

\noindent \textbf{(b) Perfect alignment.} We provide perfect alignment between the mixed image $M$ and transmission $T$. Existing datasets~\cite{wan2017benchmarking,zhang2018single} has the misalignment issue between $M$ and $T$. The misalignment issue does not only degrade a model's performance but also makes evaluation less accurate. A reflection removal model trained on misaligned paired data often generate blurry images~\cite{wei2019single_ERR}. Note that a single pixel shift in an image can affect evaluation metrics PSNR and SSIM significantly.

\noindent \textbf{(c) Diversity.} We provide much more diverse data by utilizing different types of glasses exsiting in our daily life, including colored and curved glasses shown in Fig.~\ref{fig:DiverseData}. We capture various objects in different environments and lighting conditions with three different types of cameras. As mentioned before, when we capture images in the wild, we also guarantee diversity of the smoothness and intensity of reflection.

% The type of reflection and transmission is mainly decided by glass properties in physics. Although previous methods~\cite{zhang2018single,wei2019single_ERR,wan2017benchmarking} collected data in the wild, they only use one or three glasses. While they can get different reflection by adjusting the focus (e.g., the reflection is blurry if it is out of focus), the diversity is limited. 
%We also separate the dataset according to the characteristic of reflection, which can help us to analyze the bottleneck of existing algorithms. 

\noindent \textbf{(d) Large-scale data.} Our dataset is significantly larger than existing datasets or benchmarks. In total, our dataset contains 1,063 triplets ${M,R,T}$.% Compared with SIR$^2$~\cite{wan2017benchmarking}, our dataset is large enough to provide training data for deep learning based algorithms. 
%It can provide more fair evaluation since the domain gap problem is well addressed. 
We believe our evaluation is more accurate because there is no misalignment between the mixed image and the ground truth.

\noindent \textbf{(e) Other advantages.} Compared with existing datasets~\cite{zhang2018single,wei2019single_ERR,Li_2020_CVPR}, we provide reflection image since many works have demonstrate the effectiveness of using reflection~\cite{eccv18refrmv_BDN,Lei_2020_CVPR}. We also provide raw images instead of only RGB images for future study in reflection removal. 
% As shown in Fig. \ref{fig:MMR_sample}, the $M-R$ operation must be implemented in the raw image space to achieve good performance. 
% Moreover, we observe that raw data has many advantages compared with RGB data (Sec. 4.2). 
% We also provide raw images to facilitate research in reflection removal. 

Our CDR dataset is publicly available, which can be used for training and evaluation. The detailed categories can help researchers understand the strengths and weaknesses of existing methods. We hope it can accelerate the research in reflection removal.

\begin{figure*}[t]
\centering
\includegraphics[width=.8\linewidth]{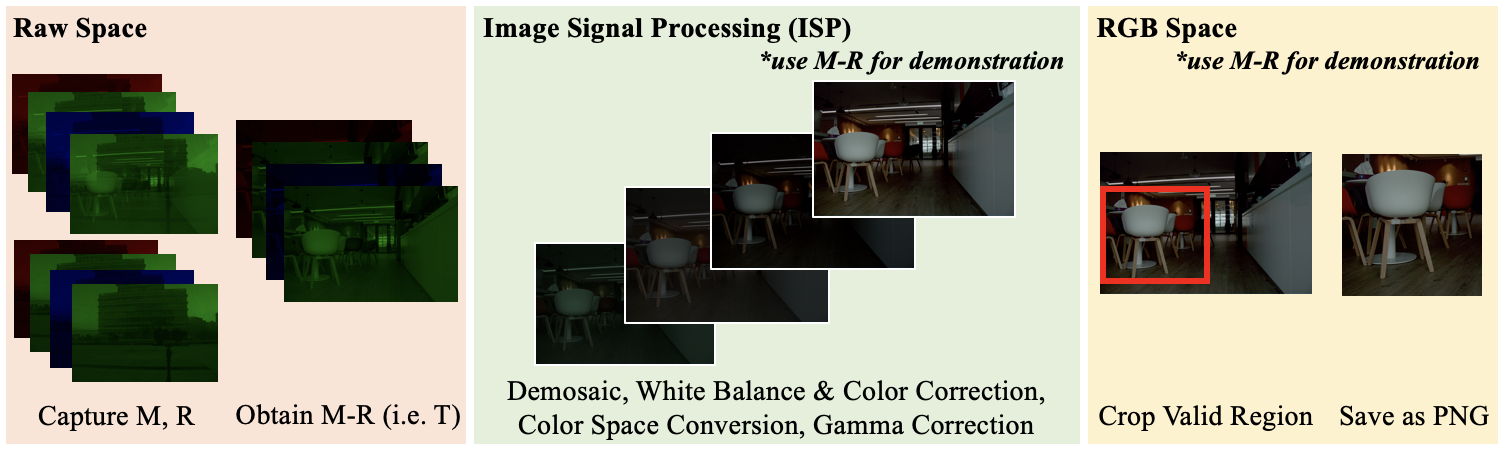}
%\caption{We need to convert raw data to RGB images.}
\caption{The post-processing pipeline. Ground-truth transmission $T$ is obtained in the RAW space. Then all the RAW images are passed through an ``ISP'' to obtain the corresponding RGB images. Finally, the regions of interest regions are cropped out.}
\label{fig:Postprocessing}
\end{figure*}

\begin{figure*}[t]
\centering
\begin{tabular}{@{}c@{\hspace{1mm}}c@{\hspace{1mm}}c@{\hspace{1mm}}c@{\hspace{1mm}}c@{\hspace{1mm}}c@{}}

% \rotatebox{90}{\small \hspace{4mm}After ISP}
\includegraphics[width=0.19\linewidth]{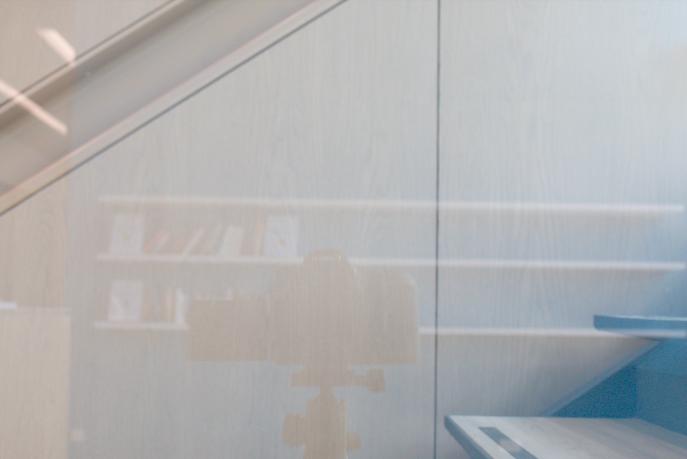}&
\includegraphics[width=0.19\linewidth]{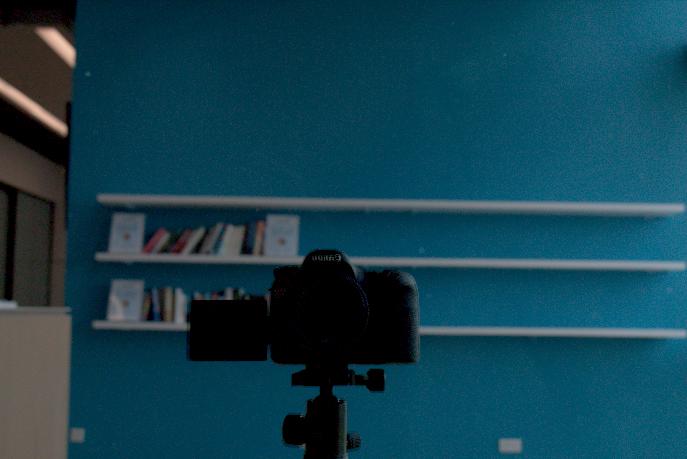}&
\includegraphics[width=0.19\linewidth]{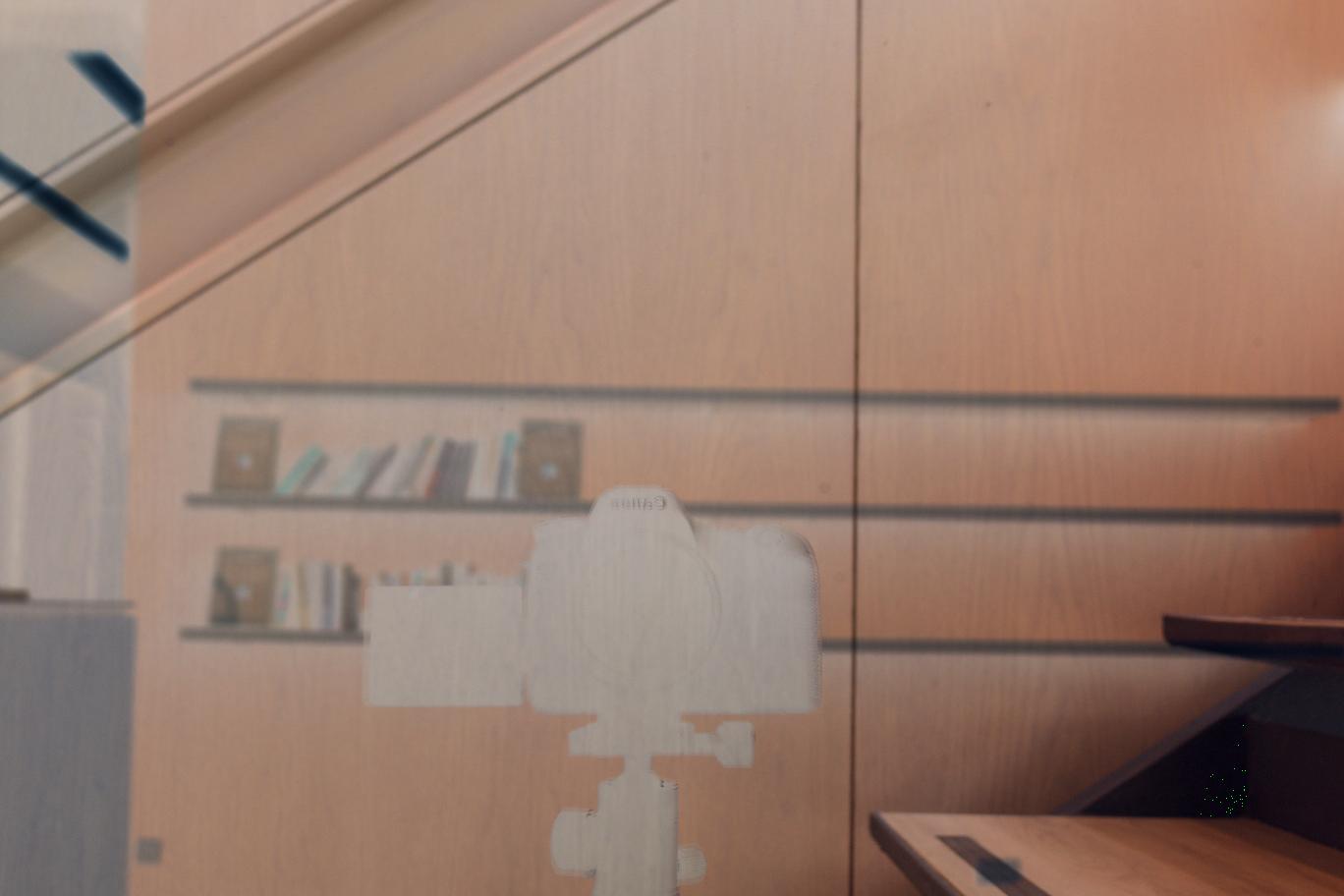}&
\includegraphics[width=0.19\linewidth]{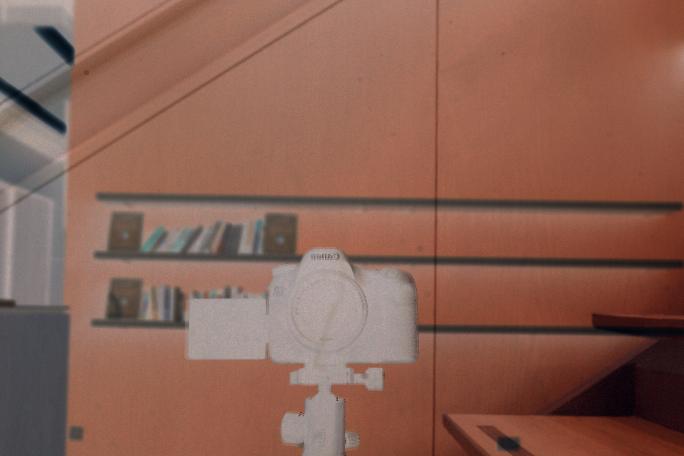}&
\includegraphics[width=0.19\linewidth]{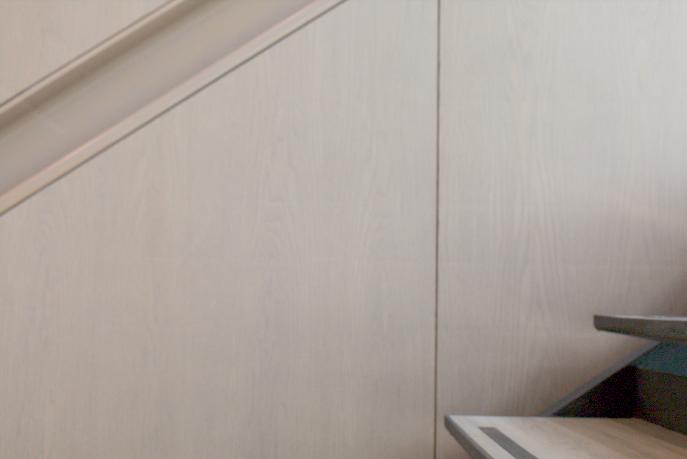}\\

RGB $M$ & RGB $R$ & RGB $M-R$ & Gamma $M-R$ & Raw $M-R$\\
\end{tabular}
% \vspace{-0.1in}
\caption{If $M-R$ is applied other than the raw data space, undesirable residuals will appear. ``RGB $M-R$'': do $M-R$ on RGB images. ``Gamma $M-R$'': use $M^{2.2}-R^{2.2}$ to reduce the impact of gamma correction. ``Raw $M-R$'': do $M-R$ on raw data.}
\label{fig:MMR_sample}
\end{figure*}

\subsection{Data Aquisition}

In order to collect diverse data with perfect alignment in the wild, we adopt the M-R pipeline proposed by Lei et al.~\cite{Lei_2020_CVPR} where $T$ is obtained by $M-R$. Different from Lei et al.~\cite{Lei_2020_CVPR} which implements $M-R$ on raw data obtained by a polarization sensor, we use normal cameras that provide the raw data to construct the CDR dataset. %We choose the M-R pipeline because the $T$ obtained by this pipeline has a perfect alignment, which is quite important to both training and evaluation. 

Fig. \ref{fig:DataCollection} shows our data collection pipeline. The first step is to have an appropriate glass. Since we do not need to remove the glass to obtain background $B$ as other methods do~\cite{wan2017benchmarking,wei2019single_ERR,zhang2018single}, we can utilize immovable glasses in the real world. As the second step, we use a piece of black cloth behind the glass to block transmission to obtain the reflection $R$. In the end, we remove the cloth to collect the mixed image $M$. Please refer to Fig.~\ref{fig:DiverseData} for some captured example iamges. %collect dynamic data in transmission. To achieve this goal, we only need to take multiple mixed images M$_1$, M$_2$, M$_3$, ..., M$_n$ after removing the black cloth. In this case, only one $R$ is collected so we require the reflection $R$ should be static here, 

There are some details in real capturing progress:

\begin{itemize}
\renewcommand{\labelitemi}{\textbullet}
    \item To ensure perfect alignment between $M$ and $R$, we use a tripod to fix the camera to take images.
    \item To ensure the exposure is consistent between $M$ and $R$, all the cameras are set to the manual mode with a fixed camera setting including ISO and exposure time.
    \item To reduce the noise level in $M$ and $R$ as much as possible, we capture data with a long exposure time and a small ISO.
    \item The objects in the reflection (not transmission) need be static in both $M$ and $R$ to ensure the perfect alignment in $M-R$.
\end{itemize}

\subsection{Post Processing}
Fig. \ref{fig:Postprocessing} shows the overall pipeline of our post-processing step. With raw $M$ and raw $R$, we calculate raw $T$ by $T=M-R$ in the raw data space. Note that $M-R$ should be implemented in the RGB space because the linearity between light intensities and RGB values does not hold, as shown in Fig.~\ref{fig:MMR_sample}. In $T=M-R$, negative values may appear due to noise, and they are set to zeros directly. The black level of a camera is added back to ensure its ISP can be applied to $T$. So far, the raw data format of $T$ is the same as $M$ and $R$. 

Since most existing reflection removal algorithms adopt RGB images as input, we need to convert raw images from the raw data space to RGB space. However, the camera default ISP is not public. Therefore, we implement our own image signal processing (ISP) pipeline to generate RGB images for $M$, $T$, and $R$ using the same ISP. As for the metadata of $T$, we simply apply the metadata of $M$ to $T$ directly. 
% Using the same ISP of $M$ can avoid the algorithm to learn the new ISP of $T$. 

% In Fig.~\ref{fig:ISP_Adobe}, we compare the performance of our ISP with a professional ISP software Lightroom and the camera default ISP. 
% Various kinds of ISP can be applied to raw $T$, and many of them can achieve satisfied perceptual results, as shown in Fig.~\ref{fig:ISP_Adobe}. In other words, it is necessary to obtain an ISP for RGB $T$ if we want to obtain a perfect result from RGB $M$. 
 
\begin{figure}[t]
\centering

\begin{tabular}{@{}c@{\hspace{1mm}}c@{\hspace{1mm}}c@{}}
\includegraphics[width=0.32\linewidth]{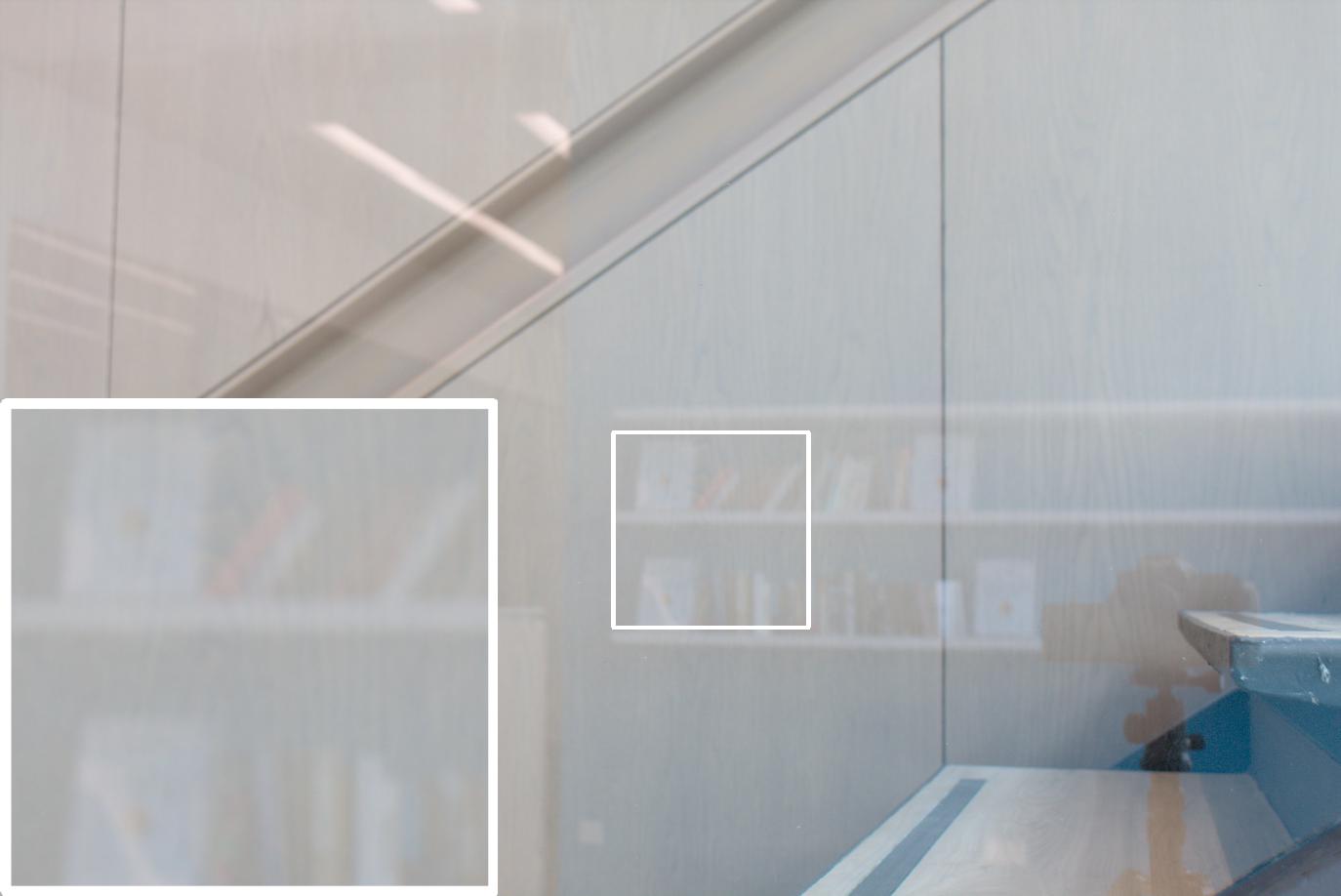}&
\includegraphics[width=0.32\linewidth]{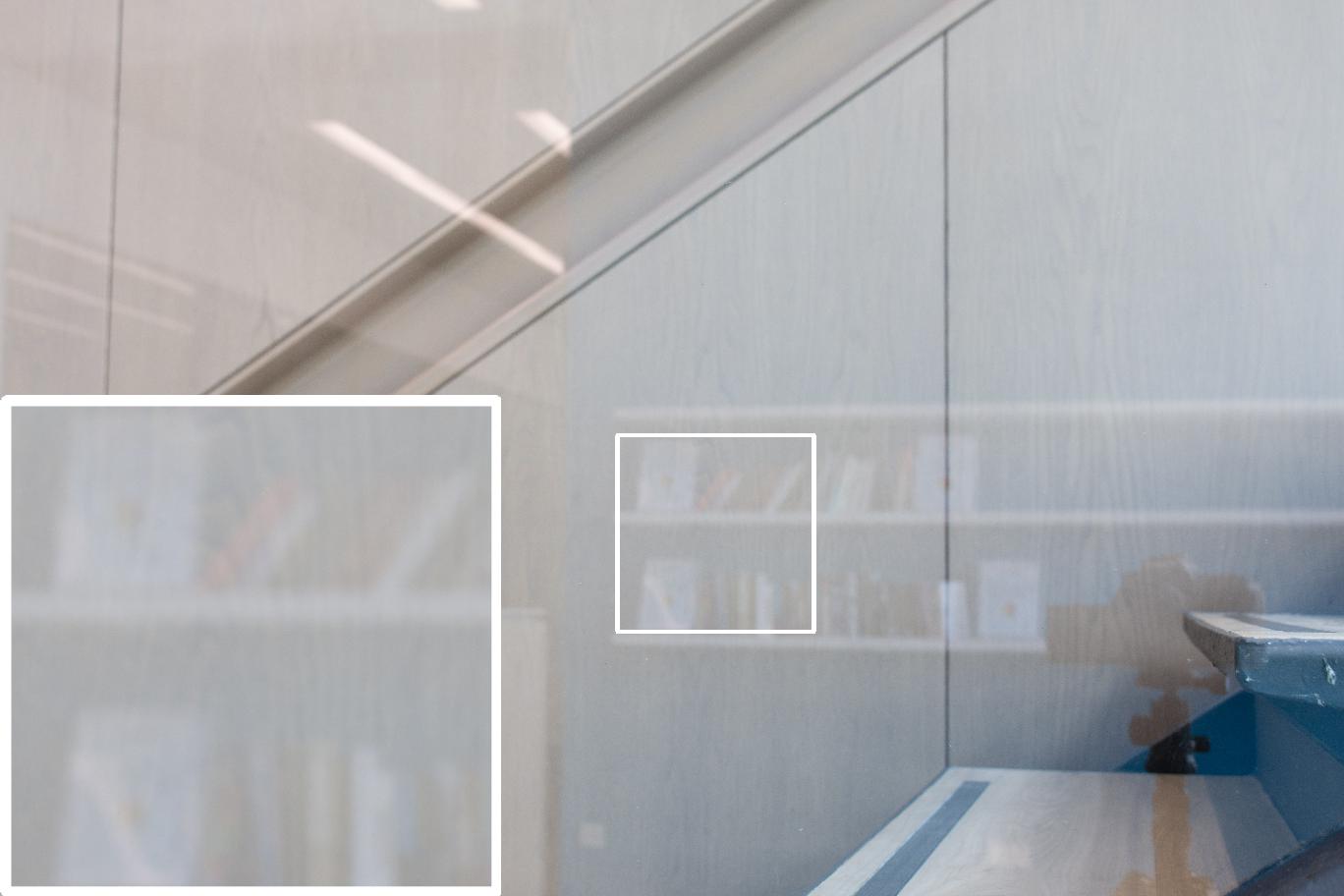}&
\includegraphics[width=0.32\linewidth]{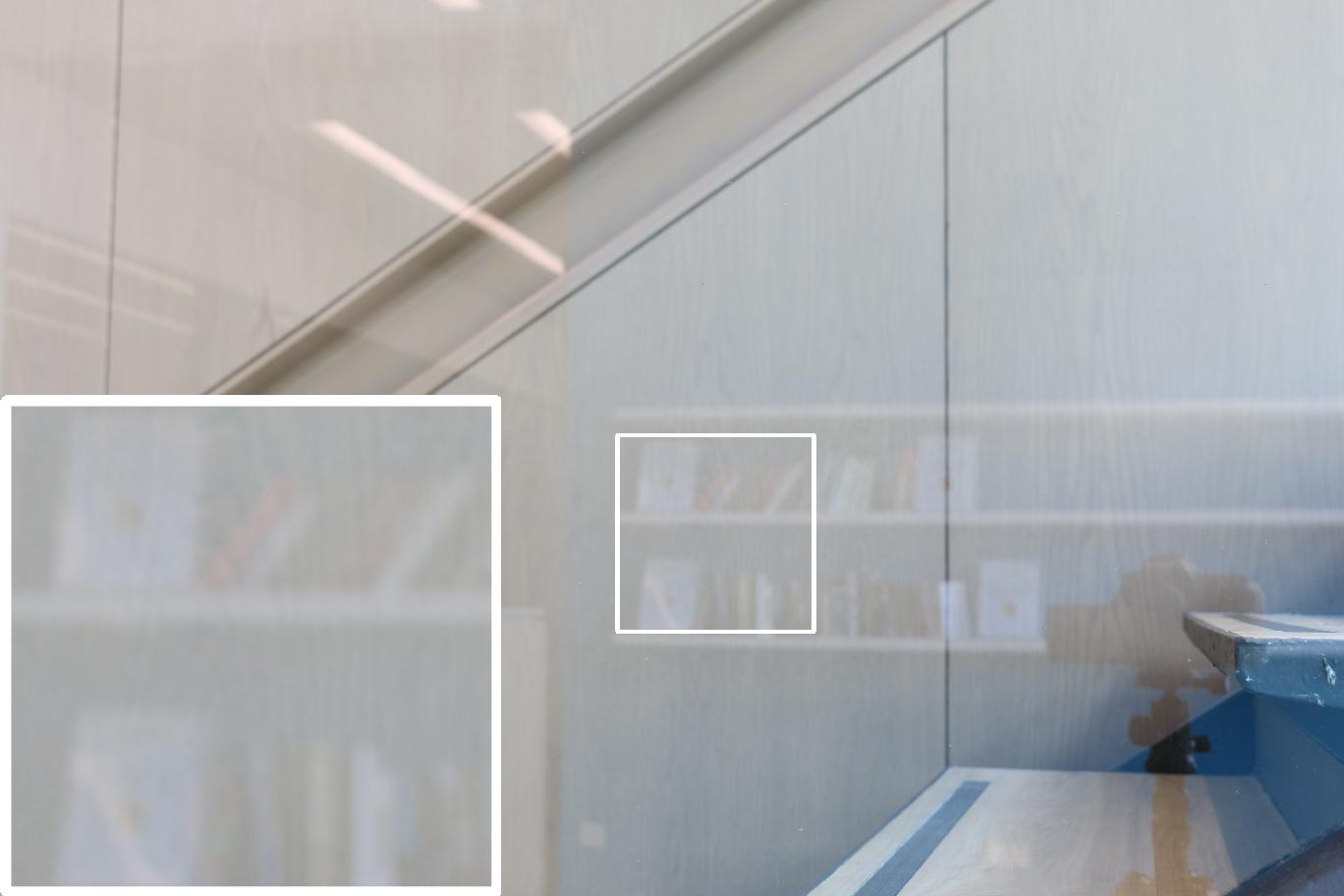}\\
(a) $M$: Our ISP & (b) $M$: Lightroom & (c) $M$: Camera\\
\includegraphics[width=0.32\linewidth]{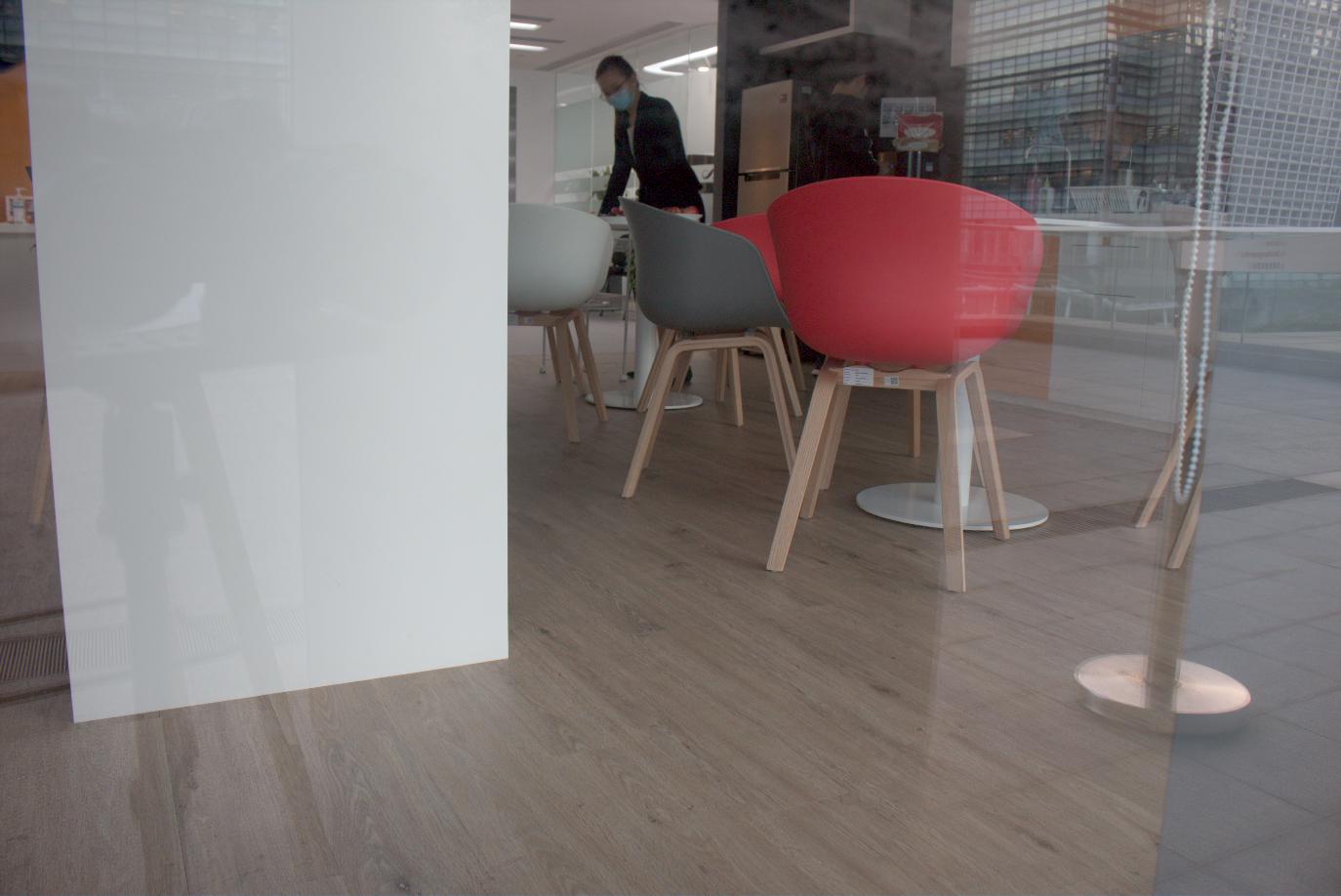}&
\includegraphics[width=0.32\linewidth]{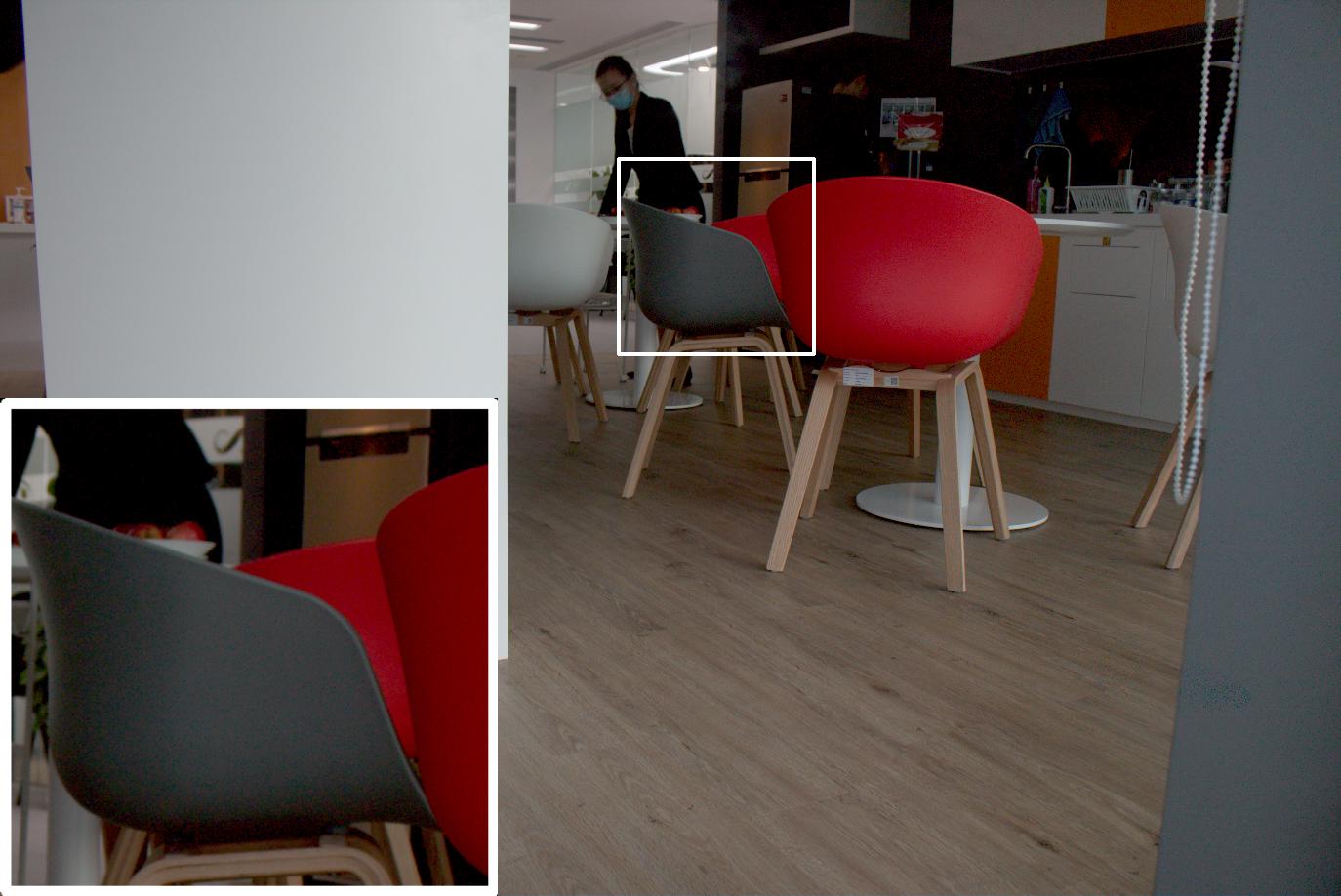}&
\includegraphics[width=0.32\linewidth]{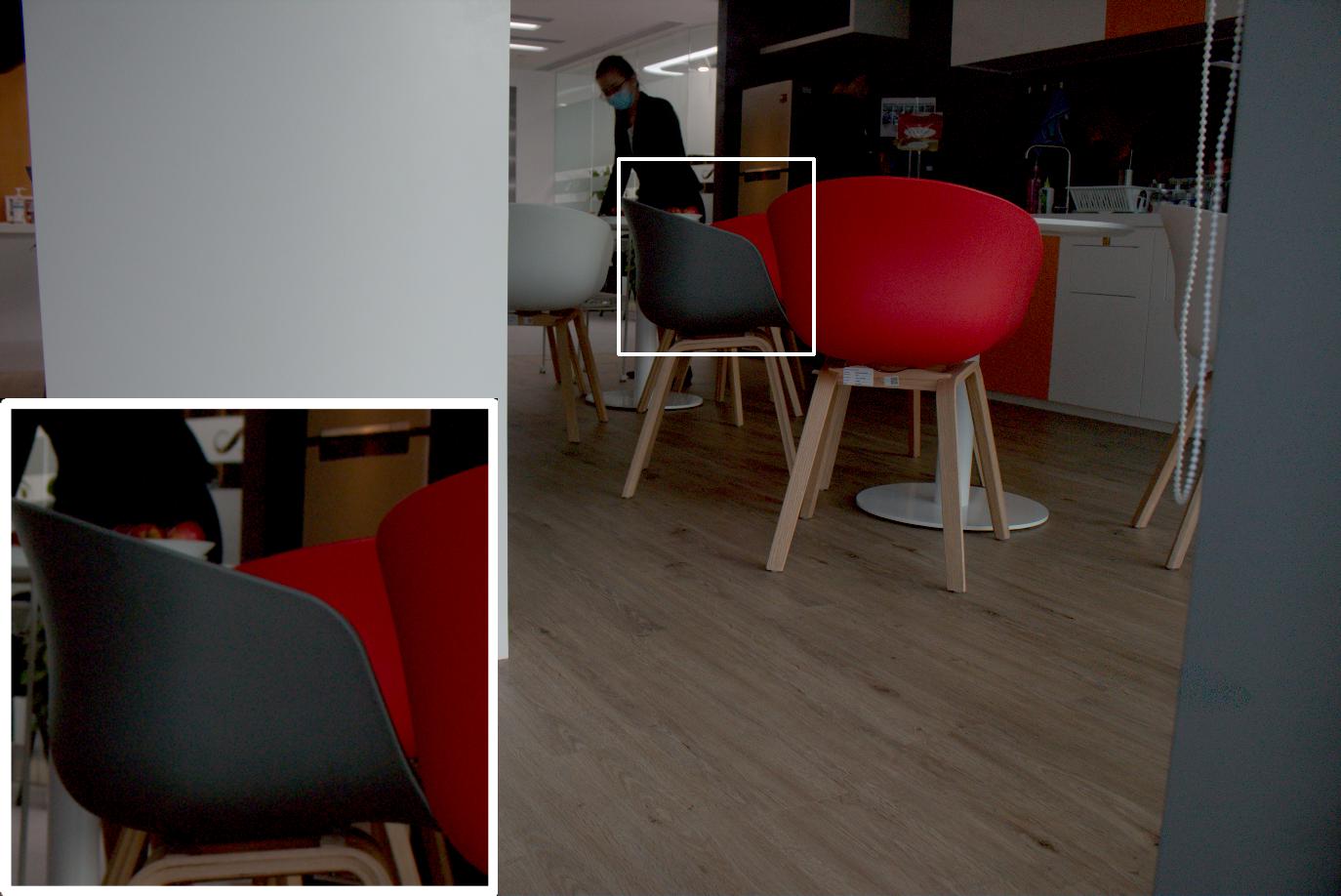}\\
(d) $M$: Our ISP1 & (e) $T$: Our ISP1 & (f) $T$: Our ISP2 \\

\end{tabular}
\caption{The first row shows that our ISP generates similar results compared with Lightroom and Camera Output. The second row shows that different ISPs can be applied to $T$ to achieve similar results. Note that Lightroom is a professional ISP software supported by Adobe.}%Note that we cannot use Lightroom or Camera's ISP for M-R since the format of M-R is not correct.
\label{fig:ISP_Adobe}
\end{figure}

After obtaining RGB $T$ images, we need to crop the area of interest for $T$. In the dataset, we eliminate the triplets  in the following two cases: (1) If multiple glasses exist behind the covered glass (i.e., glasses exist in the background), the obtained $T$ still has glasses. (2) If there is no glass in $M$, then no black cloth exists to cover the transmission. In this case, the obtained $T$ will equal 0, which contradicts the ground truth. 

Different from SIR$^2$ that classify images as bright or dark scenes according to absolute intensity, focus, or thickness of glass, we categorize images according to three criteria: relative intensity, smoothness between $R$ and $T$ and the ghosting effect. First, we observe that the impact of reflection is decided by the relative intensity instead of absolute intensity. We calculate the mean intensity ratio for each pair of $R$ and $T$ and categorize them as weak reflection, moderate reflection or strong reflection. Second, we sort the data based on the smoothness of reflection and transmission: BRST (blurry reflection and sharp transmission), SRST (sharp reflection and sharp transmission), BRBT (blurry reflection and blurry transmission). In fact, the BRST type has been generally used in previous work. At last, we pick up the data which has ghosting effect as a class.%Also, the SRST type is focused on recent work~\cite{Wen_2019_CVPR_Linear}. 

%In the day, the reflection is apparent from outdoor to indoor. In the night, the reflection is apparent from indoor to outdoor. Although the light of indoor does not change, the impact of reflection is changed due to the difference in outdoor light, which influences the relative intensity.

\section{Experiments}
%The detailed information of our CDR dataset is described in Table~\ref{table:Dataset comparison}. 
%We select high-quality data from each camera and each reflection type for evaluation. 
We first describe our experimental setup. The images from all three cameras are provided for training, which can reduce the impacts of the domain gap. Following previous work~\cite{wan2017benchmarking}, we choose PSNR, SSIM and NCC as our main evaluation metrics. %We also adopt a new metric PNCC~\cite{Lei_2020_CVPR} as a reference, which is calculated by summation of NCC on different feature maps in a pretrained VGG-19 network~\cite{vgg}. 
%The height and width of images lie in the range of [400, 3000] and [400, 3200], respectively. In the training process, images are randomly cropped into $224\times224$ to train all the methods. 
%We tried using SI and NCC which is used in SIR$^2$. However, the performance of these two metrics seems cannot represent the real performance. Due to limited space, only PSNR, SSIM and PNCC are presented. 

% \begin{align}
%     L_{PNCC}(I_A, I_B)=
%     &\sum_{l=1}^n{NCC(v_l(\tilde I_A), v_l(\tilde I_B))
%     },
% \end{align}
% where $v_l$ denotes the $l$-th layer feature maps of VGG-19 \cite{simonyan2014very}. In practice, we use three layers 'conv2\_2','conv3\_2','conv4\_2'. PNCC can also be applied using other pre-trained neural network.

The baselines for comparison in our experiment include CEILNet~\cite{fan2017generic}, CoRRN~\cite{CoRRN}, BDN~\cite{eccv18refrmv_BDN}, Zhang et al.~\cite{zhang2018single},  Wei et al.~\cite{wei2019single_ERR}, Yang et al.~\cite{Yang_2019_CVPR}, Arvanitopoulos et al.~\cite{Arvanitopoulos_2017_CVPR}, Li et al.~\cite{li2014single}, IBCLN~\cite{Li_2020_CVPR}, and Kim et al.~\cite{Kim_2020_CVPR}. All these methods take a single RGB image as input. For learning-based methods, we use their pre-trained models by default as some methods do not provide training code. 
%For Wen et al.~\cite{Wen_2019_CVPR_Linear}, three possible results based on different assumptions are provided, and we choose their best result for each image for comparisons.
For Yang et al.~\cite{Yang_2019_CVPR}, different thresholds might result in different output images. Therefore, except for the original threshold, multiple thresholds are used for comparison, and the best result is reported. %In addition, models will also be retrained or finetuned if the training code is available. and limited quantitative results in ghosting images 

\subsection{Evaluation}
\label{sec:evaluation_results}

%%%%%%%%%%%%%%%%%%%%%%%%%%%%%%%%%%%%%%%%%%%%%%%%%%%
% This table is the TABLE V from our journal paper%
%%%%%%%%%%%%%%%%%%%%%%%%%%%%%%%%%%%%%%%%%%%%%%%%%%%

\begin{table*}[t]
% \small
\centering
\renewcommand{\arraystretch}{1.2}
\begin{tabular}{lcccccccccccc}
\hline
& \multicolumn{3}{c}{All} & \multicolumn{3}{c}{SRST}& \multicolumn{3}{c}{BRST}& \multicolumn{3}{c}{Non-ghosting}\\
               & {\small PSNR} & {\small SSIM} & {\small NCC}& {\small PSNR} & {\small SSIM} & {\small NCC}& {\small PSNR} & {\small SSIM} & {\small NCC}& {\small PSNR} & {\small SSIM} & {\small NCC}\\ \hline
% PSNR & \small{22.15} & \small{22.15}  & \small{22.15} & \small{22.15} & \small{22.15} & \small{22.15} & \small{22.15} & \small{22.15} & \small{22.15} & \small{22.15} & \small{22.15} & \small{22.15} \\ 
\small{Li et al.~\cite{li2014single}} & 
\small{12.73} & \small{0.650} & \small{0.721} & \small{12.26} & 
\small{0.565} & \small{0.644} & \small{13.19} & \small{0.723} & 
\small{0.789} & \small{12.56} & \small{0.624} & \small{0.703}  \\

\small{Arvan. et al.~\cite{Arvanitopoulos_2017_CVPR}} & 
\small{19.63} & \small{0.753} & \textbf{\small{0.788}} & 
\small{18.24} & \underline{\small{0.680}} & \textbf{\small{0.691}} & 
\small{20.91} & \small{0.816} &  \textbf{\underline{\small{0.873}}} & 
\small{19.00} & \small{0.727} & \textbf{\small{0.752}}  \\

\small{Yang et al.~\cite{Yang_2019_CVPR}} & 
\small{19.42} & \textbf{\underline{\small{0.767}}} & \textbf{\underline{\small{0.782}}} & \small{18.10} & 
\underline{\small{0.680}} & \textbf{\underline{\small{0.676}}} & {\small{20.65}} & \textbf{\underline{\small{0.841}}} & 
\textbf{\small{0.874}} & \small{18.78} & \textbf{\underline{\small{0.738}}} & \textbf{\underline{\small{0.744}}}  \\

\hline

\small{CEILNet~\cite{fan2017generic}} & 
\small{17.96} & \small{0.708} & \small{0.757} & \small{16.17} & 
\small{0.596} & \small{0.654} & \small{19.49} & \small{0.802} & 
\small{0.847} & \small{17.24} & \small{0.673} & {\small{0.720}}  \\

\small{Zhang et al.~\cite{zhang2018single}} & 
\small{15.20} & \small{0.694} & \small{0.703} & \small{13.52} & 
\small{0.590} & \small{0.612} & \small{16.58} & \small{0.780} & 
\small{0.785} & \small{14.48} & \small{0.662} & \small{0.677}  \\

\small{BDN~\cite{eccv18refrmv_BDN}} & 
\small{18.97} & {\small{0.758}} & \small{0.745} & \small{19.04} & 
\textbf{\small{0.713}} & \small{0.642} & \small{19.06} & \small{0.799} & 
\small{0.836} & \small{18.62} & \small{0.733} & \small{0.698}  \\

% \small{Wen et al.~\cite{Wen_2019_CVPR_Linear}} & 
% \small{20.07} & \small{0.825} & \small{0.722} & \small{17.57} & 
% \small{0.746} & \small{0.611} & \small{21.72} & \small{0.869} & 
% \small{0.799} & \textbf{\underline{\small{18.51}}} & \small{0.836} & \small{0.611}  \\

\small{Wei et al.~\cite{wei2019single_ERR}} & 
\textbf{{\small{21.01}}} & \small{0.762} & \small{0.756} & \textbf{\underline{\small{19.52}}} & 
\small{0.672} & \small{0.631} & \textbf{\underline{\small{22.36}}} & \underline{\small{0.839}}  & 
\small{0.864} & \textbf{\small{20.50}} & \small{0.731} & \small{0.713}  \\

\small{CoRRN~\cite{CoRRN}} & 
\underline{\small{20.22}} & \textbf{\small{0.774}} & \small{0.764} & \textbf{\small{20.32}} & 
\textbf{\underline{\small{0.699}}} & \underline{\small{0.656}} & \small{20.08} & \small{0.838} & 
\small{0.859} & \underline{\small{20.37}} & \textbf{\small{0.750}} & \small{0.723}  \\

\small{IBCLN~\cite{Li_2020_CVPR}} & 
\small{19.85} & \underline{\small{0.764}} & \small{0.735} & \small{18.33} & 
\small{0.671} & \small{0.613} & \underline{\small{21.14}} & \textbf{\small{0.842}} & 
\small{0.846} & \small{19.23} & \underline{\small{0.735}} & \small{0.687}  \\

\small{Kim et al.~\cite{Kim_2020_CVPR}} & 
\textbf{\underline{\small{21.00}}} & \small{0.760} & \underline{\small{0.769}} & \underline{\small{19.27}} & 
\small{0.676} & \small{0.654} & \textbf{\small{22.61}} & \small{0.833} & 
\underline{\small{0.871}} & \underline{\textbf{\small{20.42}}} & \small{0.731} & \underline{\small{0.726}}  \\

\hline

\end{tabular}

\vspace{1mm}

\begin{tabular}{lcccccccccccc}
\hline
& \multicolumn{3}{c}{Weak $R$} & \multicolumn{3}{c}{Moderate $R$}& \multicolumn{3}{c}{Strong $R$}& \multicolumn{3}{c}{Ghosting}\\
               & {\small PSNR} & {\small SSIM} & {\small NCC}& {\small PSNR} & {\small SSIM} & {\small NCC}& {\small PSNR} & {\small SSIM} & {\small NCC}& {\small PSNR} & {\small SSIM} & {\small NCC}\\ \hline
% PSNR & \small{22.15} & \small{22.15}  & \small{22.15} & \small{22.15} & \small{22.15} & \small{22.15} & \small{22.15} & \small{22.15} & \small{22.15} & \small{22.15} & \small{22.15} & \small{22.15} \\ 
\small{Li et al.~\cite{li2014single}} & 
\small{14.36} & \small{0.779} & \small{0.841} & 
\small{12.47} & \small{0.636} & \small{0.709} & 
\small{8.89} & \small{0.309} & \small{0.401} & 
\small{13.36} & \small{0.742} & \small{0.785}  \\

\small{Arvan. et al.~\cite{Arvanitopoulos_2017_CVPR}} & 
\underline{\small{23.52}} & \small{0.878} & \textbf{\underline{\small{0.941}}} & 
\small{18.43} & \small{0.744} & \textbf{\small{0.765}} & 
\small{13.56} & \small{0.397} &  \textbf{\underline{\small{0.423}}} & 
\small{21.88} & \small{0.844} & \textbf{\underline{\small{0.919}}}  \\

\small{Yang et al.~\cite{Yang_2019_CVPR}} & 
\small{23.18} & \textbf{\small{0.903}} & \underline{\small{0.937}} 
& \small{18.28} & \textbf{\underline{\small{0.754}}} & \textbf{\underline{\small{0.755}}} & 
\small{13.50} & \small{0.402} &  \textbf{\small{0.425}} & 
{\small{21.72}} & \textbf{\underline{\small{0.870}}} & \underline{\small{0.917}}  \\
\hline

\small{CEILNet~\cite{fan2017generic}} & 
\small{21.34} & \small{0.862} & \small{0.910} & \small{17.02} & 
\small{0.685} & \small{0.731} & \small{12.06} & \small{0.341} & 
\small{0.397} & \small{20.51} & \small{0.836} & \small{0.886}  \\

\small{Zhang et al.~\cite{zhang2018single}} & 
\small{17.20} &   \small{0.827} & \small{0.822} & \small{15.10}  & 
\small{0.685} & \small{0.688} & \small{9.33} &\small{0.311}  & 
\small{0.402} & \small{17.81} & \small{0.806 } & \small{0.797}  \\

\small{BDN~\cite{eccv18refrmv_BDN}} & 
\small{21.10} & \small{0.867} & \small{0.909} & \small{18.25} & 
\small{0.746} & \small{0.711} & \textbf{\underline{\small{16.15}}} & \textbf{\small{0.485}}  & 
\small{0.411} & \small{20.20} & \small{0.850} & \small{0.909}  \\

% \small{Wen et al.~\cite{Wen_2019_CVPR_Linear}} & 
% \underline{\small{22.40}} & \small{0.882} & \small{0.816} & \small{18.39} & 
% \small{0.802} & \small{0.657} & \small{14.31} & \small{0.577} & 
% \small{0.498} & \small{21.35} & \small{0.844} & \small{0.775}  \\

\small{Wei et al.~\cite{wei2019single_ERR}} & 
\textbf{\underline{\small{24.89}}}& \textbf{\underline{\small{0.901}}} & \small{0.929} & \underline{\small{19.42}}  & 
\small{0.737} & \small{0.714} & \textbf{\small{17.00}}  & \textbf{\underline{\small{0.450}}} & 
\textbf{\underline{\small{0.423}}} & \textbf{\underline{\small{22.80}}} & \textbf{\small{0.871}} & \small{0.908} \\

\small{CoRRN~\cite{CoRRN}} & 
\small{20.50} & \small{0.890}  & \small{0.928} &\textbf{\small{21.01}}  & 
\textbf{\small{0.768}} & \underline{\small{0.735}} & \small{15.12} & \underline{\small{0.433}} & 
\small{0.394} & \small{19.70} & \small{0.861}  & \small{0.911}  \\

\small{IBCLN~\cite{Li_2020_CVPR}} & 
\small{23.17} & \underline{\small{0.899}} & \small{0.908} & \small{18.98} & 
\underline{\small{0.752}} & \small{0.714} & \small{13.81} & \small{0.395} & 
\small{0.290} & \underline{\small{22.07}} & \underline{\small{0.867}} & \small{0.906}  \\

\small{Kim et al.~\cite{Kim_2020_CVPR}} & 
\textbf{\small{25.03}} & \small{0.897} & \textbf{\small{0.946}} & \textbf{\underline{\small{19.66}}} & 
\small{0.740} & \small{0.730} & \underline{\small{15.25}} & \small{0.431} & 
\small{0.416} & \textbf{\small{23.10}} & \small{0.865} & \textbf{\small{0.925}}  \\

\hline

\end{tabular}

\vspace{1mm}
\caption{Quantitative results for different methods on our dataset. A detailed analysis is presented in the paper. The first, second, and third best results are marked by bold font, bold font with underline, and underline only.}

\label{table:MainComparison}
\end{table*}

One interesting observation is that the performance of most existing single image reflection removal methods is highly related to the type of reflection.

\textbf{Quantitative results.} Table~\ref{table:MainComparison} shows the performance of the evaluated methods in terms of PSNR, SSIM, and NCC. We also split data according to smoothness, relative intensity, and ghosting effect of reflection and report separate results to analyze the impact of different image patterns. We find that learning free methods~\cite{Arvanitopoulos_2017_CVPR,Yang_2019_CVPR} achieve good performance in cases that follow their model assumptions. Although these methods usually rank poorly on average, they can perform quite well when the reflection is blurry, weak, or has ghosting effect.

In the following, we analyze the impact of different factors on the results.

\textbf{1) Impact of real data.} The methods~\cite{wei2019single_ERR,CoRRN} trained on real data achieve better performance on general real-scene data. Since Kim et al.~\cite{Kim_2020_CVPR} synthesized physically-based data for training, they also achieve good performance on real-world data. These data include SRST data, reflection with moderate intensity or strong intensity. As we can see, these methods~\cite{Kim_2020_CVPR,wei2019single_ERR,CoRRN} are often the first, second, and third best-performing methods. %The method CoRRN~\cite{CoRRN} utilizes real reflection images for training achieves the best performance. Another evidence is that the performance of Wei et al.~\cite{wei2019single_ERR} is not as good as reported in their paper because their evaluation is mainly on synthetic data.

\textbf{2) The smoothness of reflection/transmission.} The results on different smoothness of reflection and transmission are consistent with previously adopted assumptions. All methods perform relatively well the BRST (blurry reflection and sharp transmission) set. However, when the assumption does not hold (e.g., on SRST set), the performance degrades heavily. This phenomenon occurs to all evaluated algorithms. Note that in our dataset, almost 50\% percents of $M$ are SRST.%The smoothness of reflection and transmission is completely independent of the intensity. As mentioned in the previous section, most algorithms make an assumption that the transmission is in focus, and reflection is defocused.

\textbf{3) Ghosting effect.} Another assumption about reflection is the ghosting effect. Although most deep learning methods do not synthesize ghosting reflection data, they still achieve better performance on such kind of data.

\textbf{4) Reflection Intensity.} The difficulty of reflection removal increases significantly when the intensity of the reflection increases. However, the importance of reflection removal also increases in this case. When reflection is too weak, we even do not need to remove it because it is almost invisible. When the reflection is moderate or strong, the image quality suffers heavily.

%The ranking of PSNR and SSIM does not show the same trend for different algorithms. Our discussion will be focused on the SSIM-based ranking as PSNR is quite sensitive to the absolute values of images. Although absolute intensity is also important, it is more important the remove the structure of reflection in this task. From the SSIM results, we can see Wei et al.~\cite{wei2019single_ERR} achieves the best performance among benchmarked algorithms. An interesting phenomenon is that many algorithms cannot outperform the performance of input, which is also reported in~\cite{Wen_2019_CVPR_Linear}. A possible reason is that these algorithms also remove some structure in the transmission, resulting in bad PSNR and SSIM. 

% To check this suppose, the results for different reflection types are also provided. We calculate the relative difference between the result of algorithm and input to analyze the impact of reflection type. For most benchmarked algorithms, the BRST (blurry reflection and sharp transmission) achieve the best performance, which meets the assumption. However, even in this case, the performance is still not as good as the performance on synthetic data. As we can see, the performance of all existing SOTA algorithms is not really good on these real-world data. The performance is also quite different from the data collected in controlled scene in~\cite{wan2017benchmarking}. This might be a little strange because the results on synthetic are already quite satisfying. 

\begin{figure*}[t]
\centering
\begin{tabular}{@{}c@{\hspace{1mm}}c@{\hspace{1mm}}c@{\hspace{1mm}}c@{\hspace{1mm}}c@{}}

\rotatebox{90}{\small \hspace{10mm} BRST }&
\includegraphics[width=0.236\linewidth]{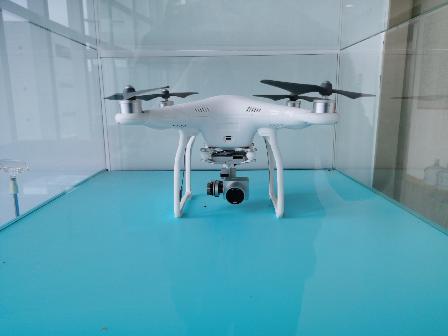}&
\includegraphics[width=0.236\linewidth]{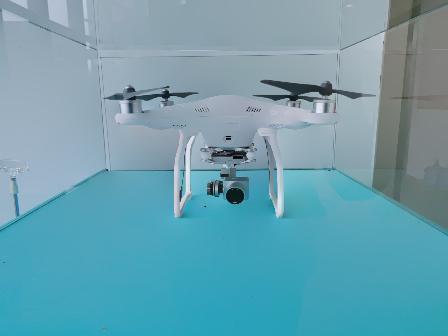}&
\includegraphics[width=0.236\linewidth]{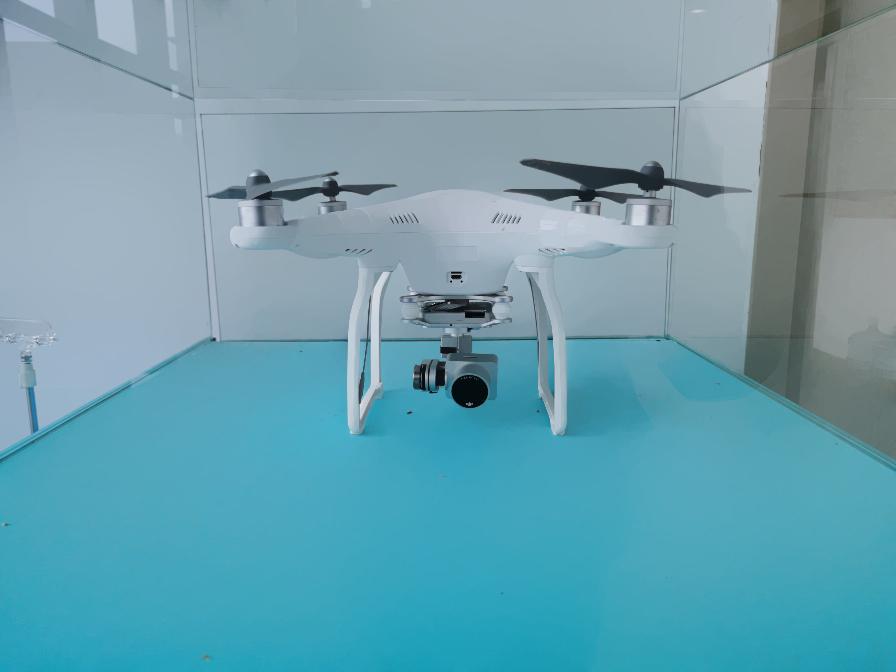}&
\includegraphics[width=0.236\linewidth]{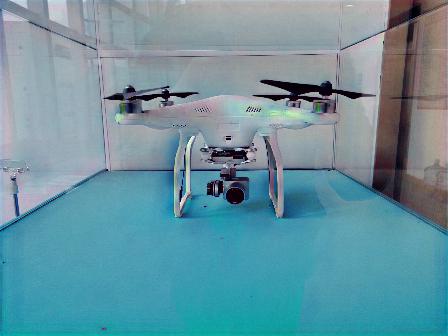}\\
&\small{Input} & \small{Yang et al.~\cite{Yang_2019_CVPR}} & \small{Arvanitopoulos et al.~\cite{Arvanitopoulos_2017_CVPR}} & \small{CEILNet~\cite{fan2017generic}}\\

&
\includegraphics[width=0.236\linewidth]{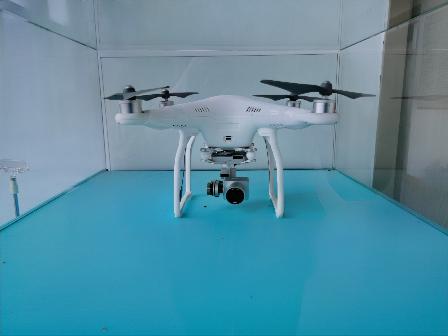}&
\includegraphics[width=0.236\linewidth]{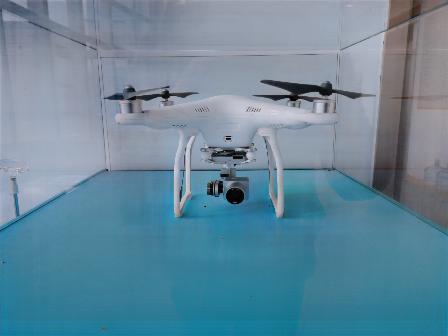}&
\includegraphics[width=0.236\linewidth]{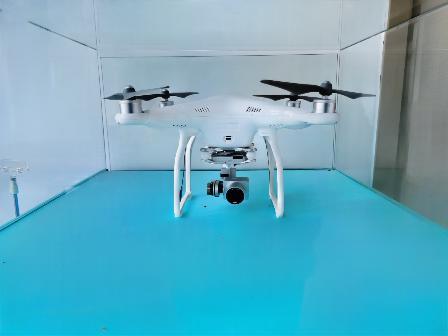}&
\includegraphics[width=0.236\linewidth]{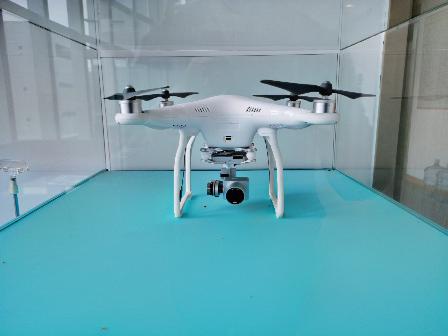}\\

&\small{CoRRN~\cite{CoRRN}} & \small{Zhang et al.~\cite{zhang2018single}} & \small{BDN~\cite{eccv18refrmv_BDN}}  & \small{Wei et al.~\cite{wei2019single_ERR} }\\

\rotatebox{90}{\small \hspace{10mm} SRST }&
\includegraphics[width=0.236\linewidth]{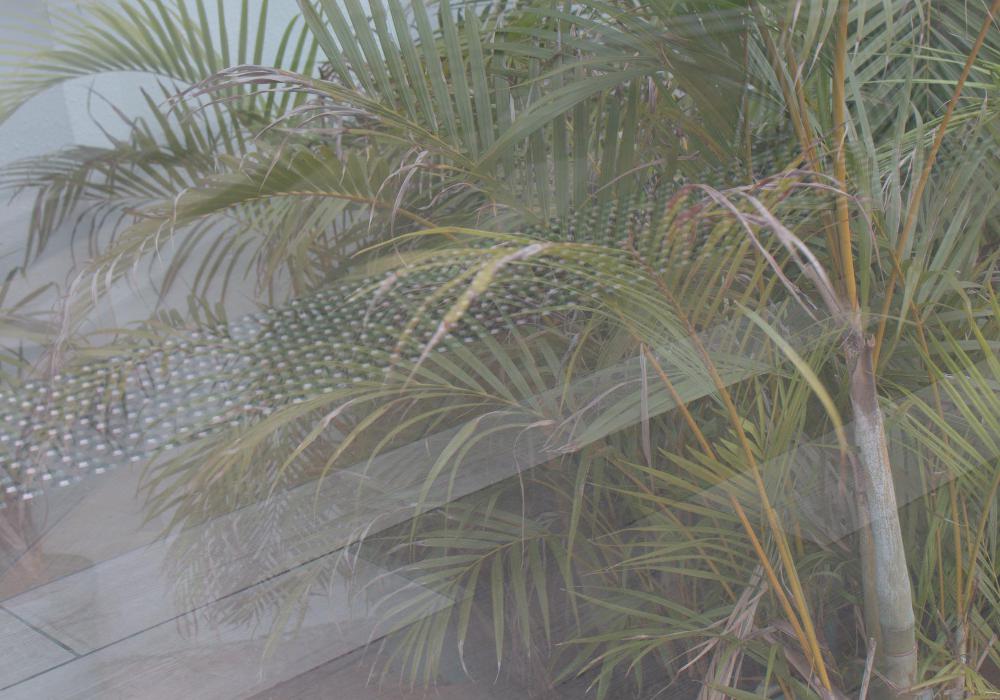}&
\includegraphics[width=0.236\linewidth]{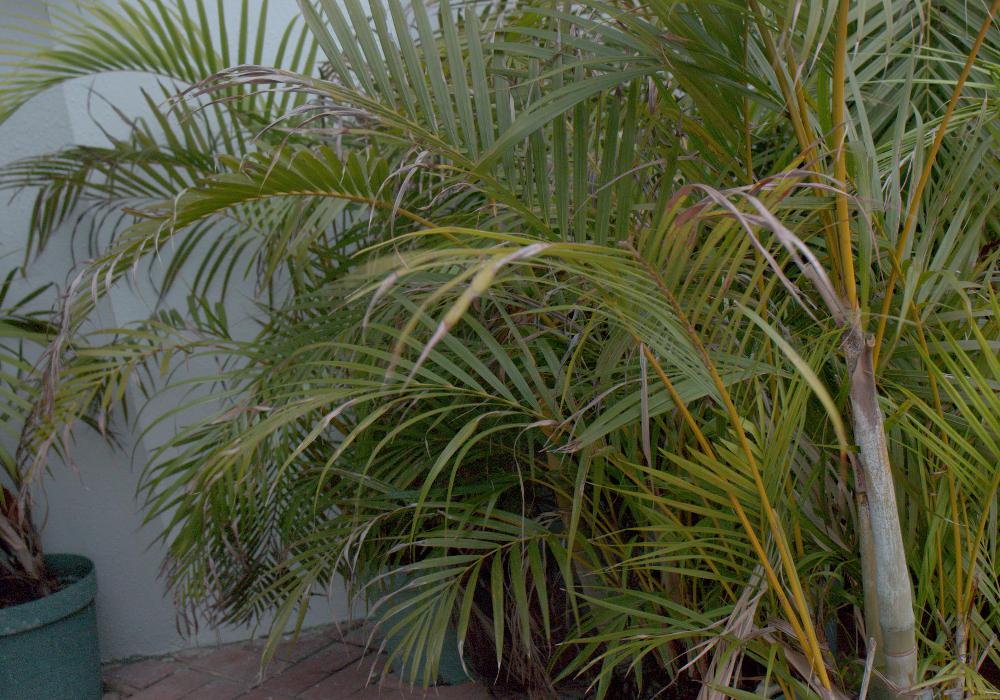}&
\includegraphics[width=0.236\linewidth]{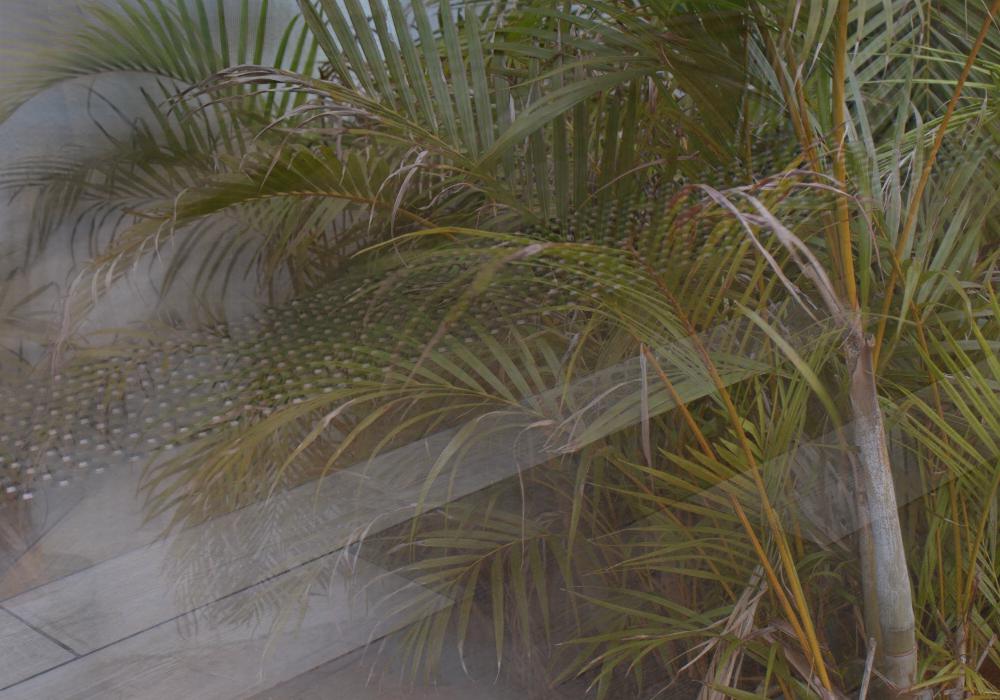}&
\includegraphics[width=0.236\linewidth]{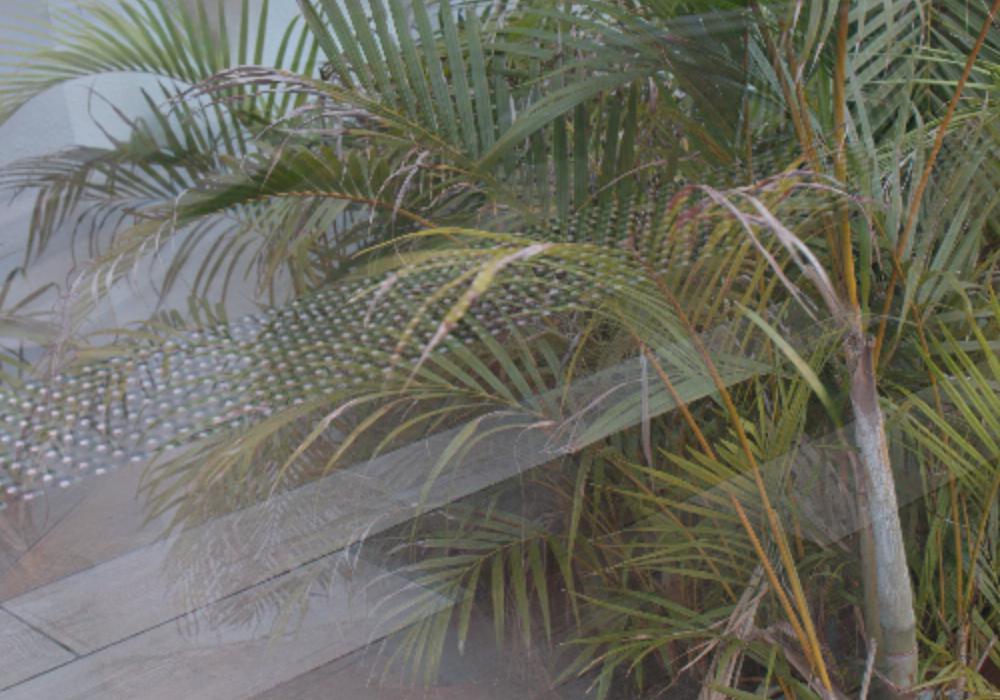}\\
&\small{Input} & \small{GT} & \small{Zhang et al.~\cite{zhang2018single}} & \small{Wei et al.~\cite{wei2019single_ERR} }\\
&
\includegraphics[width=0.236\linewidth]{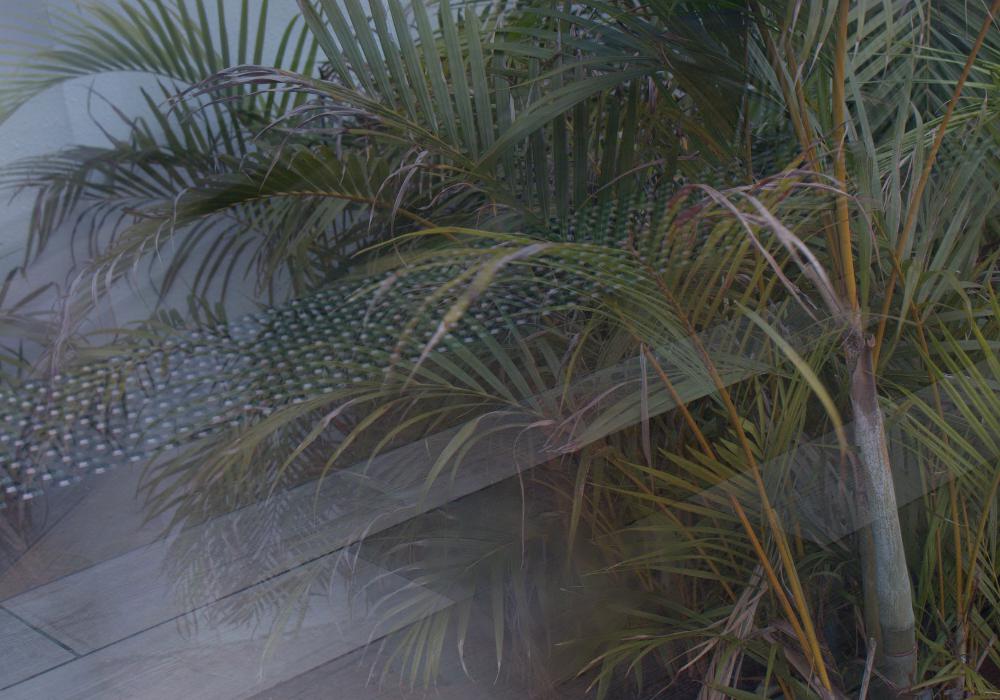}&
\includegraphics[width=0.236\linewidth]{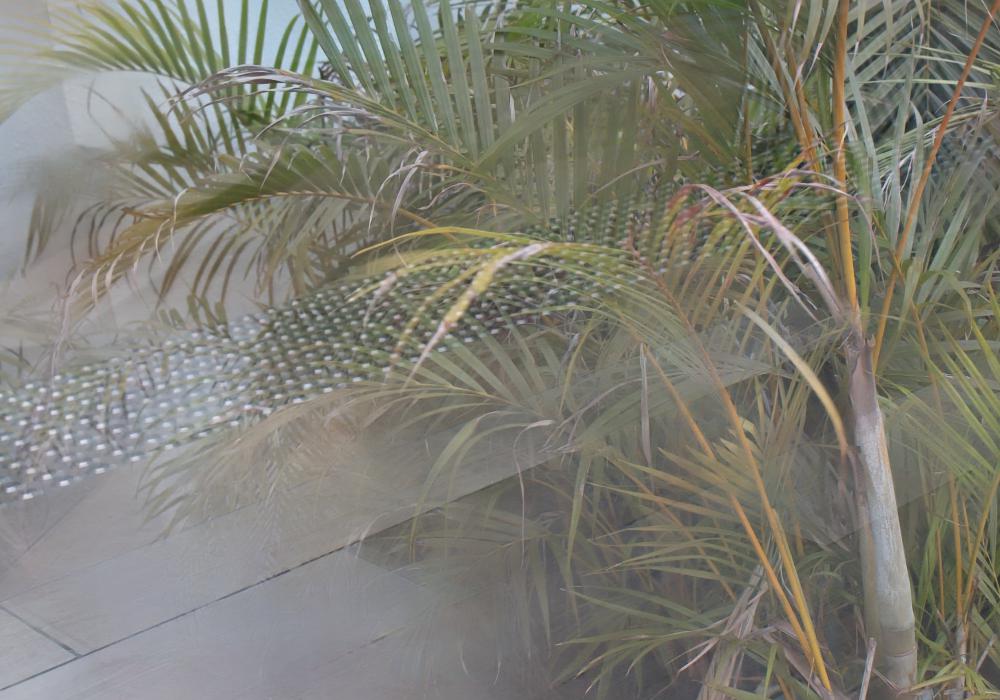}&
\includegraphics[width=0.236\linewidth]{Figure/Experiments/PerceptualCmp/C_8681_8680_Input.jpg}&
\includegraphics[width=0.236\linewidth]{Figure/Experiments/PerceptualCmp/C_8681_8680_Input.jpg}\\
&\small{CoRRN~\cite{CoRRN}} & \small{BDN~\cite{eccv18refrmv_BDN}} & \small{Yang et al.~\cite{Yang_2019_CVPR}}  & \small{Li et al.~\cite{li2014single}}
\end{tabular}
% \vspace{-0.1in}
\caption{Most methods cannot remove the sharp reflection. This is probably because learning-based methods are trained on synthetic data where $R$ is blurry. Learning free methods often assume reflection is blurry. However, sharp reflection is quite common in the real world. Figure best viewed in the electronic version.}
\label{fig:Perceptual Comparison}
% \vspace{-0.1in}
\end{figure*}

\textbf{Qualitative results.}
We present qualitative results in Fig.~\ref{fig:Perceptual Comparison} to analyze the results further. The perceptual performance is consistent with the quantitative results on different reflection types.

The performance on SRST is not satisfactory. For SRST, most methods cannot remove most reflection. %This phenomenon is especially clear when the reflection is strong. %IBCLN~\cite{Li_2020_CVPR} has a similar observersation but they also cannot handle the strong reflection case.
For BRST, most benchmarked methods can remove the reflection when the reflection is weak. Learning free methods achieve good performance in this case~\cite{Yang_2019_CVPR,Arvanitopoulos_2017_CVPR}. However, it is quite rare when the reflection is both blurry and weak. %When the reflection is blurry and strong, the difference in the performance of algorithms is determined by the residual of reflection. 
%Intuitively, even when a method ``knows'' what the reflection is, they cannot completely remove the reflection. 

Another problem that occurs commonly is the degradation of image quality as shown in Fig.~\ref{fig:Perceptual Comparison}. In some cases, a method may remove the reflection nearly completely, but transmission is modified at the same time.

\subsection{Open Problems and Discussion}
From the quantitative evaluation and perceptual results, we find that state-of-the-art single image reflection removal methods are still far from perfect, although they have achieved great performance on synthetic data or real-world data in a controlled environment.

From our experiment, we find that evaluation on synthetic data is flawed. The improvement of results on synthetic data or real-world data in a controlled environment cannot represent real improvement. If we want to apply the reflection removal method in our daily life, we should aim to achieve excellent performance on real-world data collected in the wild.

We believe we should relax strong assumptions in reflection removal methods. Strong assumptions may make the method perform well on a certain type of reflection but fail on other types. As analyzed above, most methods can achieve satisfactory results on blurry reflection but have poor performance on sharp reflection. It is definitely a difficult task to distinguish between the sharp reflection and transmission. %On the other hand, polarization can be a strong cue but it requires specific devices. It would be promising to propose a strong model for sharp reflection on a single RGB image.

\subsection{Reflection Removal on Raw Images}
It is unclear whether single image reflection removal can benefit from raw images. However, we keep the raw data in our dataset since applying raw images has achieved amazing results on low-level computer vision tasks, including low-light image enhancement ~\cite{Chen_2018_CVPR}, super-resolution~\cite{Xu_2019_CVPR}, image denoising~\cite{Zamir2020CycleISP} and ISP~\cite{ouyang21simulator,xing21invertible}. We leave the study for raw images on single image reflection removal to future work. To the best of our knowledge, our dataset is the first dataset containing raw images for single image reflection removal.

\section{Conclusion}
In this work, we propose a new dataset CDR for single image reflection removal. Compared with other reflection removal datasets, our dataset is categorized according to reflection types, has the perfect alignment, and contains diverse scenes. 
We carefully categorize the captured images into different classes and analyze the performance of state-of-the-art methods. The experimental results show that the performance of these state-of-the-art methods is highly related to the appearance and intensity of reflection. When the pre-adopted assumptions do no hold on real-world images, the methods based on these assumptions cannot achieve top performance. We believe researchers can utilize our benchmark to do research on real-world data in the wild. In addition to RGB images, the raw data is also provided for future study.

{\small
\bibliographystyle{ieee_fullname}
\bibliography{egbib}

\begin{thebibliography}{10}\itemsep=-1pt

\bibitem{DBLP:journals/tog/AgrawalRNL05}
Amit Agrawal, Ramesh Raskar, Shree~K. Nayar, and Yuanzhen Li.
\newblock Removing photography artifacts using gradient projection and
  flash-exposure sampling.
\newblock {\em TOG}, 2005.

\bibitem{alayrac2019visual}
Jean-Baptiste Alayrac, Joao Carreira, and Andrew Zisserman.
\newblock The visual centrifuge: Model-free layered video representations.
\newblock In {\em CVPR}, 2019.

\bibitem{Arvanitopoulos_2017_CVPR}
Nikolaos Arvanitopoulos, Radhakrishna Achanta, and Sabine Susstrunk.
\newblock Single image reflection suppression.
\newblock In {\em CVPR}, 2017.

\bibitem{Chen_2018_CVPR}
Chen Chen, Qifeng Chen, Jia Xu, and Vladlen Koltun.
\newblock Learning to see in the dark.
\newblock In {\em CVPR}, 2018.

\bibitem{fan2017generic}
Qingnan Fan, Jiaolong Yang, Gang Hua, Baoquan Chen, and David Wipf.
\newblock A generic deep architecture for single image reflection removal and
  image smoothing.
\newblock In {\em ICCV}, 2017.

\bibitem{Fraid1999}
H. {Farid} and E.~H. {Adelson}.
\newblock Separating reflections and lighting using independent components
  analysis.
\newblock In {\em CVPR}, 1999.

\bibitem{DoubleDIP}
Yossi Gandelsman, Assaf Shocher, and Michal Irani.
\newblock "double-dip": Unsupervised image decomposition via coupled
  deep-image-priors.
\newblock In {\em CVPR}, 2019.

\bibitem{guo2014robust}
Xiaojie Guo, Xiaochun Cao, and Yi Ma.
\newblock Robust separation of reflection from multiple images.
\newblock In {\em CVPR}, 2014.

\bibitem{han2017reflection}
Byeong-Ju Han and Jae-Young Sim.
\newblock Reflection removal using low-rank matrix completion.
\newblock In {\em CVPR}, 2017.

\bibitem{NIR_Hong}
Y. {Hong}, Y. {Lyu}, S. {Li}, and B. {Shi}.
\newblock Near-infrared image guided reflection removal.
\newblock In {\em ICME}, 2020.

\bibitem{Kim_2020_CVPR}
Soomin Kim, Yuchi Huo, and Sung-Eui Yoon.
\newblock Single image reflection removal with physically-based training
  images.
\newblock In {\em CVPR}, 2020.

\bibitem{kong14pami}
Naejin Kong, Yu-Wing Tai, and Joseph~S. Shin.
\newblock A physically-based approach to reflection separation: from physical
  modeling to constrained optimization.
\newblock {\em TPAMI}, 2014.

\bibitem{Lei_2021_RFC}
Chenyang Lei and Qifeng Chen.
\newblock Robust reflection removal with reflection-free flash-only cues.
\newblock In {\em CVPR}, 2021.

\bibitem{Lei_2020_CVPR}
Chenyang Lei, Xuhua Huang, Mengdi Zhang, Wenxiu Sun, Qiong Yan, and Qifeng
  Chen.
\newblock Polarized reflection removal with perfect alignment in the wild.
\newblock In {\em CVPR}, 2020.

\bibitem{Li_2020_CVPR}
Chao Li, Yixiao Yang, Kun He, Stephen Lin, and John~E. Hopcroft.
\newblock Single image reflection removal through cascaded refinement.
\newblock In {\em CVPR}, 2020.

\bibitem{li2013exploiting}
Yu Li and Michael~S Brown.
\newblock Exploiting reflection change for automatic reflection removal.
\newblock In {\em ICCV}, 2013.

\bibitem{li2014single}
Yu Li and Michael~S Brown.
\newblock Single image layer separation using relative smoothness.
\newblock In {\em CVPR}, 2014.

\bibitem{Liu_CVPR_2020}
Yu-Lun Liu, Wei-Sheng Lai, Ming-Hsuan Yang, Yung-Yu Chuang, and Jia-Bin Huang.
\newblock Learning to see through obstructions.
\newblock In {\em CVPR}, 2020.

\bibitem{Lyu_2019_Polar}
Youwei Lyu, Zhaopeng Cui, Si Li, Marc Pollefeys, and Boxin Shi.
\newblock Reflection separation using a pair of unpolarized and polarized
  images.
\newblock In H. Wallach, H. Larochelle, A. Beygelzimer, F. d\textquotesingle
  Alch\'{e}-Buc, E. Fox, and R. Garnett, editors, {\em Advances in Neural
  Information Processing Systems}, volume~32. Curran Associates, Inc., 2019.

\bibitem{Ma_2019_ICCV}
Daiqian Ma, Renjie Wan, Boxin Shi, Alex~C. Kot, and Ling-Yu Duan.
\newblock Learning to jointly generate and separate reflections.
\newblock In {\em ICCV}, 2019.

\bibitem{ouyang21simulator}
Hao Ouyang, Zifan Shi, Chenyang Lei, Ka~Lung Law, and Qifeng Chen.
\newblock Neural camera simulators.
\newblock In {\em CVPR}, 2021.

\bibitem{eccv2018/Wieschollek}
Wieschollek Patrick, Gallo Orazio, Gu Jinwei, and Kautz Jan.
\newblock Separating reflection and transmission images in the wild.
\newblock In {\em ECCV}, 2018.

\bibitem{Punnappurath_2019_CVPR}
Abhijith Punnappurath and Michael~S. Brown.
\newblock Reflection removal using a dual-pixel sensor.
\newblock In {\em CVPR}, 2019.

\bibitem{sarel2004separating}
Bernard Sarel and Michal Irani.
\newblock Separating transparent layers through layer information exchange.
\newblock In {\em ECCV}, 2004.

\bibitem{Sarel2005}
Bernard Sarel and Michal Irani.
\newblock Separating transparent layers of repetitive dynamic behaviors.
\newblock In {\em ICCV}, 2005.

\bibitem{2000Schechner}
Yoav Schechner, Joseph Shamir, and Nahum Kiryati.
\newblock Polarization and statistical analysis of scenes containing a
  semireflector.
\newblock {\em Journal of the Optical Society of America. A, Optics, image
  science, and vision}, 2000.

\bibitem{sun2016automatic}
Chao Sun, Shuaicheng Liu, Taotao Yang, Bing Zeng, Zhengning Wang, and Guanghui
  Liu.
\newblock Automatic reflection removal using gradient intensity and motion
  cues.
\newblock In {\em ACM-MM}, 2016.

\bibitem{szeliski2000layer}
Richard Szeliski, Shai Avidan, and P Anandan.
\newblock Layer extraction from multiple images containing reflections and
  transparency.
\newblock In {\em CVPR}, 2000.

\bibitem{wan2017benchmarking}
Renjie Wan, Boxin Shi, Ling-Yu Duan, Ah-Hwee Tan, and Alex~C Kot.
\newblock Benchmarking single-image reflection removal algorithms.
\newblock In {\em ICCV}, 2017.

\bibitem{CoRRN}
Renjie Wan, Boxin Shi, Haoliang Li, Ling-Yu Duan, Ah-Hwee Tan, and Alex~Kot
  Chichung.
\newblock Corrn: Cooperative reflection removal network.
\newblock {\em TPAMI}, 2019.

\bibitem{wang2015automatic}
Qiaosong Wang, Haiting Lin, Yi Ma, Sing~Bing Kang, and Jingyi Yu.
\newblock Automatic layer separation using light field imaging.
\newblock {\em arXiv preprint arXiv:1506.04721}, 2015.

\bibitem{wei2019single_ERR}
Kaixuan Wei, Jiaolong Yang, Ying Fu, David Wipf, and Hua Huang.
\newblock Single image reflection removal exploiting misaligned training data
  and network enhancements.
\newblock In {\em CVPR}, 2019.

\bibitem{Wen_2019_CVPR_Linear}
Qiang Wen, Yinjie Tan, Jing Qin, Wenxi Liu, Guoqiang Han, and Shengfeng He.
\newblock Single image reflection removal beyond linearity.
\newblock In {\em CVPR}, 2019.

\bibitem{xing21invertible}
Yazhou Xing, Zian Qian, and Qifeng Chen.
\newblock Invertible image signal processing.
\newblock In {\em CVPR}, 2021.

\bibitem{Xu_2019_CVPR}
Xiangyu Xu, Yongrui Ma, and Wenxiu Sun.
\newblock Towards real scene super-resolution with raw images.
\newblock In {\em CVPR}, 2019.

\bibitem{xue2015computational}
Tianfan Xue, Michael Rubinstein, Ce Liu, and William~T Freeman.
\newblock A computational approach for obstruction-free photography.
\newblock {\em TOG}, 2015.

\bibitem{eccv18refrmv_BDN}
Jie Yang, Dong Gong, Lingqiao Liu, and Qinfeng Shi.
\newblock Seeing deeply and bidirectionally: a deep learning approach for
  single image reflection removal.
\newblock In {\em ECCV}, 2018.

\bibitem{Yang_2019_CVPR}
Yang Yang, Wenye Ma, Yin Zheng, Jian-Feng Cai, and Weiyu Xu.
\newblock Fast single image reflection suppression via convex optimization.
\newblock In {\em CVPR}, 2019.

\bibitem{Zamir2020CycleISP}
Syed~Waqas Zamir, Aditya Arora, Salman Khan, Munawar Hayat, Fahad~Shahbaz Khan,
  Ming-Hsuan Yang, and Ling Shao.
\newblock Cycleisp: Real image restoration via improved data synthesis.
\newblock In {\em CVPR}, 2020.

\bibitem{zhang2018single}
Xuaner Zhang, Ren Ng, and Qifeng Chen.
\newblock Single image reflection separation with perceptual losses.
\newblock In {\em CVPR}, 2018.

\end{thebibliography}
}

\end{document}